\documentclass[sigconf]{acmart}
\usepackage{booktabs} % For formal tables
\usepackage{amsmath, amsfonts, amssymb, xspace, color, amsbsy, subcaption}
\usepackage{tabularx}
\usepackage{tikz, graphicx}
\usepackage{enumitem}
\usepackage{multirow}
\usetikzlibrary{positioning}
\usepackage[linesnumbered,ruled,lined]{algorithm2e}
\usepackage[font=small,labelfont=bf]{caption} %,labelfont=bf,textfont=bf

\newcommand{\ReMine}{{\sf ReMine}\xspace}
\newcommand{\ReMineL}{{\sf ReMine-L}\xspace}
\newcommand{\ReMineG}{{\sf ReMine-G}\xspace}
\newcommand{\xhdr}[1]{\vspace{1.7mm}\noindent{{\bf #1}}}

\newcommand{\hide}[1]{} %hide

\newcommand{\etal}{\emph{et~al.}\xspace} % and others
\newcommand{\ie}{\emph{i.e.}\xspace} % that is
\newcommand{\eg}{\emph{e.g.}\xspace} % for example
\newcommand{\Sum}{\sum\limits} % Cheap displaystyle operators

\newcommand{\norm}[1]{\mathopen\| #1 \mathclose\|}			% use instead of $\|x\|$

\DeclareMathOperator*{\argmax}{argmax}

\def \e {\mathbf{e}}

\def \h {\mathbf{h}}

\def \r {\mathbf{r}}
\def \t {\mathbf{t}}

\def \P {\mathbf{P}}

\def \C {\mathcal{C}}
\def \D {\mathcal{D}}
\def \F {\mathcal{F}}

\def \O {\mathcal{O}}

\def \S {\mathcal{S}}
\def \T {\mathcal{T}}

\def \one {\mathbf{1}}

\SetKwRepeat{Do}{do}{while}

\setcopyright{rightsretained}

\setlist[itemize]{leftmargin=18pt}
\setlist[enumerate]{leftmargin=24pt}
\begin{document}
\title{Integrating Local Context and Global Cohesiveness\\ for Open Information Extraction}
\author{Qi Zhu$^1$, Xiang Ren$^2$, Jingbo Shang$^1$, Yu Zhang$^1$, Ahmed El-Kishky$^1$, Jiawei Han$^1$}

\affiliation{%
	\institution{{$^1$}{University of Illinois at Urbana-Champaign, IL, USA}}
}
\affiliation{%
	\institution{{$^2$}{University of Southern California, CA, USA}}
}
\affiliation{%
	\institution{{$^1$}{\{qiz3, shang7, yuz9, elkishk2, hanj\}@illinois.edu} $\quad$ {$^2$}{xiangren@usc.edu} }
}

%!TEX root = main.tex
\begin{abstract}
Extracting entities and their relations from text is an important task for understanding massive text corpora. Open information extraction (IE) systems mine relation tuples (i.e., entity arguments and a predicate string to describe their relation) from sentences. These relation tuples are not confined to a predefined schema for the relations of interests. However, current Open IE systems focus on modeling \textit{local} context information in a sentence to extract relation tuples, while ignoring the fact that \textit{global} statistics in a large corpus can be \textit{collectively} leveraged to identify high-quality sentence-level extractions.
In this paper, we propose a novel Open IE system, called \ReMine,  which integrates local context signals and global structural signals in a unified, distant-supervision framework. Leveraging facts from external knowledge bases as supervision, the new system can be applied to many different domains to facilitate sentence-level tuple extractions using corpus-level statistics.
%The new system can be efficiently applied to different domains as it uses facts from external knowledge bases as supervision; and can effectively score sentence-level tuple extractions based on corpus-level statistics. 
Our system operates by solving a joint optimization problem to unify (1) segmenting entity/relation phrases in individual sentences based on local context; and (2) measuring the quality of tuples extracted from individual sentences with a translating-based objective. Learning the two subtasks jointly helps correct errors produced in each subtask so that they can mutually enhance each other.
Experiments on two real-world corpora from different domains demonstrate the effectiveness, generality, and robustness of \ReMine when compared to state-of-the-art open IE systems.

\end{abstract}
%%%%%%%%%%%%%%%%%%%%%%%%%%%%%%%%%%%%%%%%%%%%%%%%%%%%%%

\copyrightyear{2019} 
\acmYear{2019} 
\setcopyright{acmcopyright}
\acmConference[WSDM '19]{The Twelfth ACM International Conference on Web Search and Data Mining}{February 11--15, 2019}{Melbourne, VIC, Australia}
\acmBooktitle{The Twelfth ACM International Conference on Web Search and Data Mining (WSDM '19), February 11--15, 2019, Melbourne, VIC, Australia}
\acmPrice{15.00}
\acmDOI{10.1145/3289600.3291030}
\acmISBN{978-1-4503-5940-5/19/02}

\settopmatter{printacmref=false}

%\keywords{Open Information Extraction}

\maketitle

%!TEX root = main.tex
\section{Introduction}
\begin{figure}[t]
	\centering
	\includegraphics[width=0.90\linewidth]{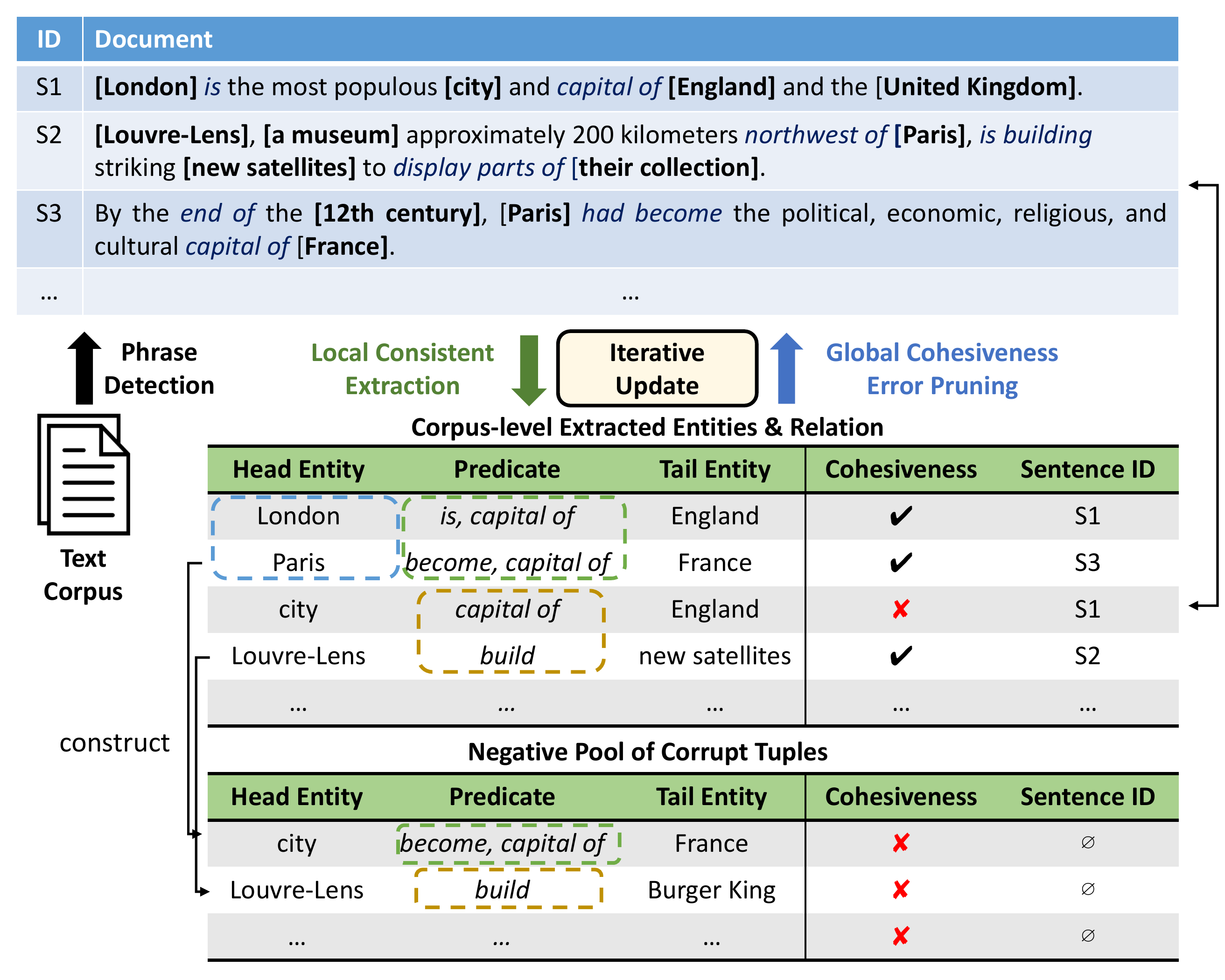}
	\vspace{-0.2cm}
	\caption{Example of incorporating global cohesiveness view for error pruning. One can infer ``London" and ``Paris" are similar because they co-occur a lot with the same relation in corpus. By constructing false tuples from extractions,  ``city'' occurs with relation ``capital of'' in the negative pool more often, then it is unlikely for tuples with ``city" and ``capital of" to be correct.}
	\label{fig:ReMine-eg}
\vspace{-0.2cm}
\end{figure}

%Massive text corpora are emerging worldwide in different domains and languages. The sheer size of such unstructured data and the rapid growth of new data pose grand challenges on making sense of these massive corpora.
With the emergence of massive text corpora in many domains and languages, the sheer size and rapid growth of this new data poses many challenges understanding and extracting insights from these massive corpora.
Information extraction (IE)~\cite{sarawagi2008information} -- extraction of relation tuples in the form of $\langle$\textit{head entity}, \textsf{relation}, \textit{tail entity}$\rangle$ -- is a key step towards automating knowledge acquisition from text. In Fig.~\ref{fig:ReMine-eg}, for example, the relation tuple $\langle$\textit{Louvre-Lens}, \textsf{build}, \textit{new satellites}$\rangle$ can be extracted from unstructured text $s_2$ to represent a piece of factual knowledge in structured form. These relation tuples have a variety of downstream applications, such as serving as building blocks for knowledge base construction~\cite{dong2014knowledge} and facilitating question answering systems~\cite{faderopenQA14,sun2015open}.
While traditional IE systems require people to pre-specify the set of relations of interest, recent studies on \textit{open-domain information extraction} (Open IE)~\cite{banko2007open,carlson2010coupled,schmitz2012open} rely on \textit{relation phrases} extracted from text to represent the entity relationship, making it possible to adapt to various domains (\ie, open-domain) and different languages (\ie, language-independent). 

\begin{figure*}[t]
\vspace{-0.4cm}
  \centering
  \includegraphics[width=0.90\textwidth]{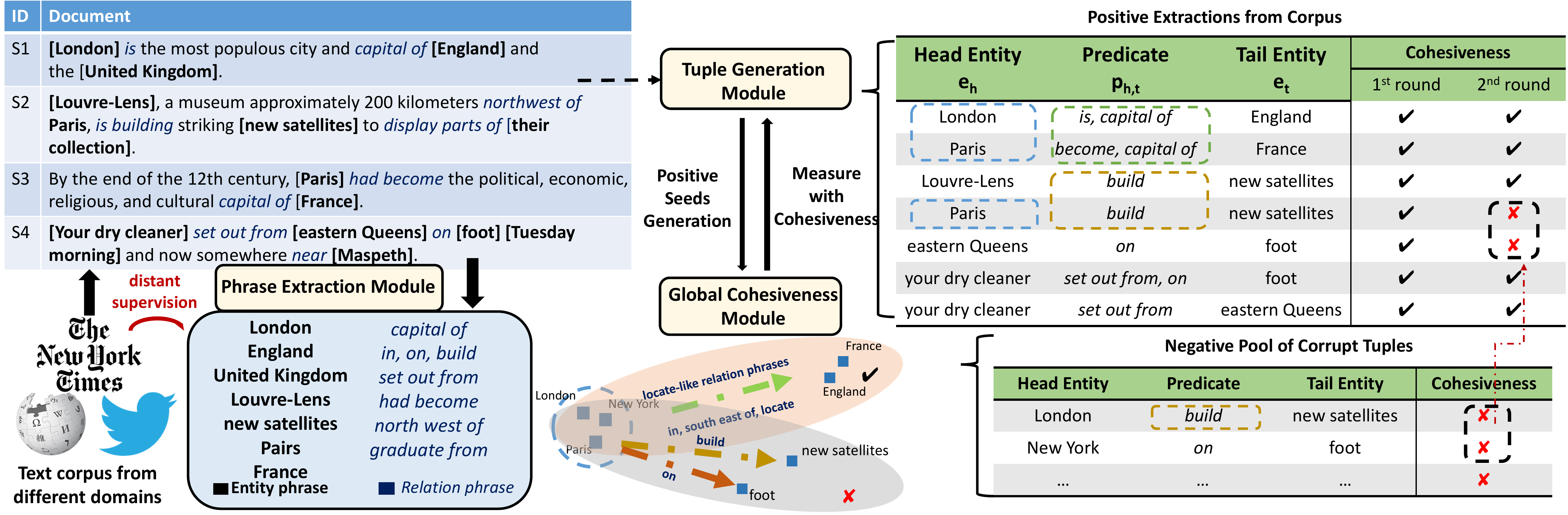}
  \vspace{-0.2cm}
  \caption{Overview of the \ReMine Framework.}
  \label{fig:Framework}
\end{figure*}

Current Open IE systems focus on analyzing the local context within individual sentences to extract entities and their relationships, while ignoring the redundant information that can be collectively referenced across different sentences and documents in the corpus. For example, in Fig.~\ref{fig:ReMine-eg}, seeing entity phrases ``\textit{London}" and ``\textit{Paris}" frequently co-occur with similar predicate strings and tail entities in the corpus, one gets to know that they have close semantics (same for ``\textit{England}" and ``\textit{France}"). This not only helps confirm that $\langle$\textit{London}, \textsf{is capital of}, \textit{England}$\rangle$ is a quality tuple as we know $\langle$\textit{Paris}, \textsf{become capital of}, \textit{France}$\rangle$ is extracted with high confidence, but this also rules out the tuple $\langle$\textit{city}, \textsf{capital of}, \textit{England}$\rangle$ as ``\textit{city}" is semantically distant from ``\textsf{capital of}".
Therefore, the information redundancy in the massive corpus provides clues on whether a candidate relation tuple is consistently used in the corpus, and motivates us to design a principled way of measuring tuple quality (\ie, global cohesiveness). 

Furthermore, most existing Open IE systems assume that they have access to entity detection tools (\eg, named entity recognizer (NER), noun phrase (NP) chunker) to extract entity phrases from sentences, which are then used to form entity pairs for relation tuple extraction~\cite{banko2007open,carlson2010coupled,schmitz2012open}. Some systems further rely on dependency parsers to generate syntax parse trees to guide the relation tuple extraction~\cite{schmitz2012open,del2013clausie,angeli2015leveraging}. However, these systems suffer from \textit{error propagation} as the errors in prior parts of the pipeline(\eg, entity recognition) could accumulate by cascading down the pipeline(\eg, to relation tuple extraction), yielding more significant errors. In addition, the NERs and NP chunkers are often pre-trained for general domain and may not work well on a domain-specific corpus (\eg, scientific papers, social media posts).

\nocite{Bunescu2007LearningTE}

In this paper, we propose a novel framework, called \ReMine, to unify two important yet \textit{complementary} signals for the Open IE problem, \ie, the local context information and global cohesiveness (see also Fig.~\ref{fig:Framework}). While most existing Open IE systems focus on analyzing local context and sentence structures for tuple extraction, \ReMine further makes use of all the candidate tuples extracted from the entire corpus, to collectively measure whether these candidate tuples are reflecting cohesive semantics. This is done by mapping both entity and relation phrases into the same low-dimensional embedding space, where two entity phrases are similar if they share similar relation phrases and head/tail entity phrases. The entity and relation embeddings so learned can be used to measure the cohesiveness score of a candidate relation tuple. To overcome the error propagation issue, \ReMine \textit{jointly} optimizes both the \textit{extraction of entity and relation phrases} and the \textit{global cohesiveness across the corpus}, each being formalized as an objective function so as to quantify the quality scores, respectively.

%Specifically, \ReMine first identifies entity and relation phrases from local context. In Fig.~\ref{fig:Framework}, suppose we have a sentence ``Your dry cleaner set out from eastern Queens on foot Tuesday morning and now somewhere near Maspeth.''. 
%We will first extract five entity phrases, \textit{eastern Queens}, \textit{Tuesday morning}, \textit{Maspeth}, \textit{Your dry cleaner}, and \textit{foot}. 
%Then, \ReMine jointly mines relation tuples and measures extraction with the global translating objective. Local-consistent text segmentation may generate noisy tuples, such as $<${your dry cleaner}, \textit{set out from}, {eastern Queens}$>$ and $<${eastern Queens}, \textit{on}, {foot}$>$. 
%However, from the global cohesiveness view, we may infer that the second tuple as a false positive. Entity phrases like ``eastern Queens'' are seldom linked by relation phrase \textit{``on''} in extracted tuples. On the opposite, similar tuple $<${New York}, \textit{on}, {foot}$>$ appears frequently in the corrupted negative pool.
%Overall, \ReMine will iteratively refine extracted tuples and learn entity and relation representations from a corpus level.
%With careful attention to advantages of linguistic patterns~\cite{hearst1992automatic,fader2011identifying} and representation learning~\cite{bordes2013translating}, this approach benefits from both sides. Additionally, compared to previous open IE systems, \ReMine prunes extracted tuples via global cohesiveness while demonstrating performance insensitive to the target domain.

The major contributions of this paper are as follows.
\begin{enumerate}[leftmargin=*,noitemsep]
\item We propose a novel open IE framework, \ReMine, that can extract relation tuples with local context and global cohesiveness.
\item We develop a context-dependent phrasal segmentation algorithm that can identify high quality phrases of multiple types.
\item We design a unified objective to measure both tuple quality in a local context and global cohesiveness of candidate tuples.
\item Extensive experiments on three public datasets demonstrate that \ReMine achieves state-of-the-art performance on both entity phrase extraction task as well as Open IE task.
%when compared with various baseline methods.
\end{enumerate}
%!TEX root = main.tex
\section{Problem Definition}
%The input of Open IE is a collection of documents $\D$, where the goal is to first identify entity phrases $\mathcal{E}$ , relation phrases $\R$, as well as extract meaningful relation tuples $\T$ among them.

%\begin{definition}
\xhdr{Notations.}
For any sentence s in a corpus $\D$, a \textsl{phrase}, \textsl{p}, is defined as single-word or multi-word phrase in s. We further group phrases into three different types, \ie \textsl{entity phrase} \textsl{e}, \textsl{relation phrase} \textsl{r} and \textsl{background text} \textsl{b}. In Open IE, an \textsl{entity phrase} occurs as subject or object in extractions. In practice, entity phrase can be either a named entity of pre-defined types(\eg, \textit{time, location, person, organization}) or other noun phrases. In sentence $s_4$ of Fig.~\ref{fig:Framework}, ``Your dry cleaner'' is not a named entity, although it is the subject of this sentence and cannot be omitted in relation tuples extraction. Therefore, previous work~\cite{fader2011identifying,schmitz2012open} use pre-trained NP chunkers to identify entity phrases. \textsl{Positive entity phrase pairs} $E_p^+$ is a set of entity pairs that may have textual relations between them. 
\textsl{Relation phrase r} describes relation between an entity phrase pair ($e_h$, $e_t$) $\in E_p^+$. Unlike relation extraction tasks, one relation instance can correspond to multiple relation phrases, \textit{\eg location/country/capital can correspond to ('s capital, capital of, the capital, ...)}. Lastly, \textsl{background text} is not a component of relation tuple.

\xhdr{Problem.}
Let $\T$ denote the extracted relation tuples. Each \textsl{relation tuple t} is defined as \{$e_h$, $p_{h,t}$, $e_t$\}, where $e_h$ and $e_t$ correspond to head and tail entity arguments and predicate $p=(r_1, r_2, ... r_n)$ may contain multiple relation phrases(\eg, we have two relation phrases: ``had become'' and ``capital of'' between $\langle$Paris, France$\rangle$ in sentence $s_3$).  Formally, we define the task of Open IE as follows.

%Different from existing Open Information Extraction systems~\cite{angeli2015leveraging,schmitz2012open}, we treat predicate as combination of relation phrase, \eg \{Your dry cleaner, foot, (\textit{set\_out, on})\}. Unlike the relation types used by external knowledge bases, predicate in relation tuples can be ambiguous and may not align with relation types. We will discuss how to capture semantic drift in one predicate and identify relation phrases $(r_1, r_2, ... r_n)$ in our formulation. Formally, we define the task of Open IE as follows.

\begin{definition}
\textsl{Given corpus $\D$, the task of Open IE \textbf{aims to}: (1) segment sentence $s \in \D$ to extract entity phrases $e$, relation phrases $r$; and (2) output relation tuples $\{e_h, p_{h,t}, e_t\}^{N_t}_{k=1}$,.}
\end{definition}

%!TEX root = main.tex
\section{The \ReMine Framework}
\ReMine aims to jointly address two problems: extracting entity and relation phrases from sentences and generating quality relation tuples. To accomplish this, we must first address three challenges. First, as the phrase boundary and category are unknown, one needs to design a segmentation algorithm to score the quality of segmented phrases and label their categories.
%the true label of phrases in the target domain is unknown, therefore scoring candidate phrases needs an effective phrase-quality measure. 
Second, as multiple entity phrases may be extracted from a sentence, one needs to identify positive entity phrase pairs and obtain proper relation phrase between them.
%there exist multiple entity phrases in one sentence. Therefore, selecting entities to form relation tuples may suffer from ambiguity in local context. 
Third, as tuple extraction based solely on local sentence context may be error-prone, one needs to incorporate corpus-level statistics to help correct errors.
%ranking extracted tuples without referring to the entire corpus may favor those with good local structures. 

\xhdr{Framework Overview.} We propose a framework, called \ReMine, that integrates both local context and global structure cohesiveness  (see also Fig.~\ref{fig:Framework}) to address above challenges. \ReMine has three major modules, each focusing on address one challenge mentioned above: (1) phrase extraction module; (2) tuple generation module; and (3) global cohesiveness module. First, to extract quality phrases with different categories, the phrase extraction module trains a robust phrase classifier using existing entity phrases from external knowledge base as ``distant supervision'' and adjust quality iteratively. Second, the tuple extraction module generates candidate tuples based on sentence's language structure---it adopts widely used local structure patterns~\cite{del2013clausie,nakashole2012patty,schmitz2012open}, including syntactic and lexical patterns over pos tags and dependency parsing tree. Different from previous studies, the module incorporates corpus-level information redundancy. Last, the global cohesiveness module learns entity and relation phrase representation and uses the  representation in a score function to rank tuples. By collaborating with each other, the relation tuple generation module and the global cohesiveness module mutually enhance each other's results. 
Particularly, the relation tuple generation module produces candidate relation tuples(as positive tuples) and feeds them into the global cohesiveness module. By distinguishing positive tuples with constructed negative samples, the global cohesiveness module provides a cohesiveness measure for candidate tuples.
The tuple generator further incorporates global cohesiveness into local generation and outputs more precise extractions. \ReMine integrates tuple generation and global cohesiveness learning into a joint objective. 
%Each module iteratively refines input for the other resulting in cleaner extractions.
Upon convergence, the training process results in distinctive and accurate tuples. Overall, \ReMine extracts relation tuples as follows, see also Fig.~\ref{fig:Framework}:
\begin{enumerate}[leftmargin=*,noitemsep]
\item \textbf{Phrase extraction module} conducts context-dependent phrasal segmentation on a target corpus (using distant supervision) , to generate entity phrases, relation phrases and sentence segmentation probability $\mathcal{W}$. 
%Apply random forest to obtain  phrase type and phrase quality score from partially labeled training data $\D_L$ (using distant supervision).
\item \textbf{Tuple generation module} generates positive entity pairs and identifies predicates $p$ between each entity phrase pair ($e_h$, $e_t$).
\item \textbf{Global cohesiveness module} learns entity and relation representations $\mathcal{V}$ via a translating objective to capture global structure cohesiveness $\sigma$.
\item \textbf{Iteratively update extractions} $\mathcal{T}$ based on both local context information and global structure cohesiveness.
\end{enumerate}

%We propose a generic information extraction framework.

\subsection{Phrase Extraction Module}
\label{sec:local_seg}
\begin{footnotesize}
\begin{example}[Multi-type phrasal segmentation] \hfill
\label{eg:phrasal_seg}
\begin{center}
\textbf{[London]} \textbf{\textit{is}} the most populous \textbf{[city]} and \textbf{\textit{captital of}} \textbf{[England]} and the \textbf{[United Kingdom]}.
\end{center}
entity phrases in \textbf{[]}, relation phrases in \textit{italic} and all the others are background text.
\end{example}
\end{footnotesize}

%\begin{example}
%In sentence $s_1$, the phrase extraction module first extracts entity phrases: \textit{London, England, city and United Kingdom}, relation phrases: \textsf{is, capital of} and all the others are either background phrases or none phrases. 
%\end{example}
%{work on this later}Traditional open IE uses NP-chunking to extract entity phrases, yet not all noun phrases can carry rich information on a target domain and thus requires additional training beyond NP-chunking. 
We address entity and relation phrase extraction as a multi-type phrasal segmentation task.
%Different from existing open information extraction methods, we realize NP-chunker mainly depend on part-of-speech tags, which could be noisy sometimes.\FX{This is probably not true. AFAIK, some np chunker are based on parse tree for more accurate results.
%Not sure if this claim is too bold.}
 Given word sequence $\C$ and corresponding linguistic features $\F$ in Table~\ref{tab:features}, a phrasal segmentation $\S = s_1, s_2, ..., s_n$ is separated by boundary index $B = b_1, b_2, ..., b_{n+1}$. For each segment $s_i$, there is a type indicator $t_i \in \{e,r,b\}$\footnote{e:entity phrase, r:relation phrase, b:background text}, indicating the most possible type of $s_i$. In above example~\ref{eg:phrasal_seg}, $s_0 = London, t_0 = e$. We factorize the phrasal segmentation probability as:
\begin{equation}
P(\mathcal{C} | \mathcal{F}) = \prod_{i=1}^{n} P(b_{i+1},s_i | b_i, \mathcal{F})
\end{equation}

\noindent
\ReMine generates each segment as follows,\\
1. Given the start index $b_i$, generate the end index $b_{i+1}$ according to context-free prior $P(b_{i+1} - b_i)$ = $\delta^{|b_{i+1} - b_i|}$, \ie length penalty~\cite{liu2015mining}.\\
%\begin{equation}
%P(b_{i+1} | b_i, \F) = \Delta(\F_{[b_i, b_{i+1})]})
%\end{equation}
2. Given the start and end index $(b_i, b_{i+1})$ of segment $s_i$, generate a word sequence $s_i$ according to a multinomial distribution over all segments of the same length.
\begin{equation}
P(s_i|b_i, b_{i+1}) = P(s_i|b_{i+1} - b_i)
\end{equation}
3. Finally, we generate a phrase type $t_i$ indicating the type of $s_i$ and a quality score showing how likely it is to be a good phrase $\lceil s \rfloor$.
\begin{equation}
P(\lceil s_i \rfloor|s_i) = \max\substack{t_i} P(t_i|s_i) = Q\substack{t_i}(s_i)
\end{equation}

%\begin{equation}
%P(b_{i+1} | b_i, \mathcal{F}) = p(\lceil{w_l...w_r}\rfloor|\mathcal{F}_{dep}) \cdot p(\lceil{w_l...w_r}\rfloor|\mathcal{F}_{punc})
%\end{equation}

%These rules have been proved lead to high recall, see Table.~\ref{tab:entity-detection}, compared with those only use frequent n-grams~\cite{2017arXiv170904109L,liu2015mining}

\begin{table}[t]
\centering
\caption{Entity and relation phrase candidates generation with regular expression patterns on part-of-speech tag}
\vspace{-1em}
\label{tab:patterns}
\begin{tabular}{l|l}
\hline
\textbf{Pattern} & \textbf{Examples}\\
\hline
\multicolumn{2}{l}{Entity Phrase Patterns}\\
\hline
\texttt{<DT|PP\$>?<JJ>*<NN>+} & \textit{the state health department}\\
\hline
\texttt{<NNP>+<IN>?<NNP>+} & \textit{Gov. Tim Pawlenty of Minnesota}\\
\hline
\multicolumn{2}{l}{Relation Phrase Patterns}\\
\hline
\texttt{\{V=<VB|VB*>+\}} & \textit{furnish}, \textit{work}, \textit{leave}\\
\hline
\texttt{\{V\}\{P=<NN|JJ|RP|PRP|DT>\}} & \textit{provided by}, \textit{retire from} \\
\hline
\texttt{\{V\}\{W=<IN|RP>?*\}\{P\}} & \textit{die peacefully at home in}\\
\hline
\end{tabular}
\end{table}

Phrase type $t$ and quality $Q$ are determined by a random forest classifier with robust positive-only distant training~\cite{shang2017automated}, which uses phrases in external knowledge base as positive samples and draws a number of phrases from unknown candidates as negative samples.
Among all word sequence $s_i$, we denote unique phrase as $u$ and $P(s_i|b_{i+1} - b_i)$ as $\theta_u$.
Similar with~\cite{liu2015mining}, we use Viterbi Training~\cite{allahverdyan2011comparative} to find best segmentation boundary $B$ and parameters $\theta, \delta$ iteratively. In the E-step, given $\theta$ and $\delta$, dynamic programming is used to find the optimized segmentation. In the M-step, we first fix parameter $\theta$, and update context-dependent prior $\delta$. Next when $\delta$ is fixed, optimized solution of $\theta_u$ is:
\begin{equation} \label{eq:update_theta}
\theta_u = \frac{\sum_{i=1}^m \one \cdot (s_i = u)}{\sum_{i=1}^m \one \cdot (b_{i+1}-b_i = |u|)}
\end{equation}
Phrase Mining~\cite{shang2017automated, book_pm} makes an assumption that quality phrases can only be frequent n-grams within a corpus. To overcome the phrasal sparsity of this assumption, several NP-chunking rules~\cite{fader2011identifying} in Table~\ref{tab:patterns}, are adopted to discover infrequent but informative phrase candidates. In experiment~\ref{exp:ner}, \ReMine has better performance than AutoPhrase~\cite{shang2017automated} as we consider more phrase candidates and multi-type segmentation helps exclude relation phrases and background text better in entity phrase extraction task.

\subsection{Tuple Generation Module}
\label{sec:tuple_generation}
\begin{figure}[t]
  \centering
	\begin{subfigure}[t]{0.49\linewidth}
	\centering
	\includegraphics[width=0.6\linewidth]{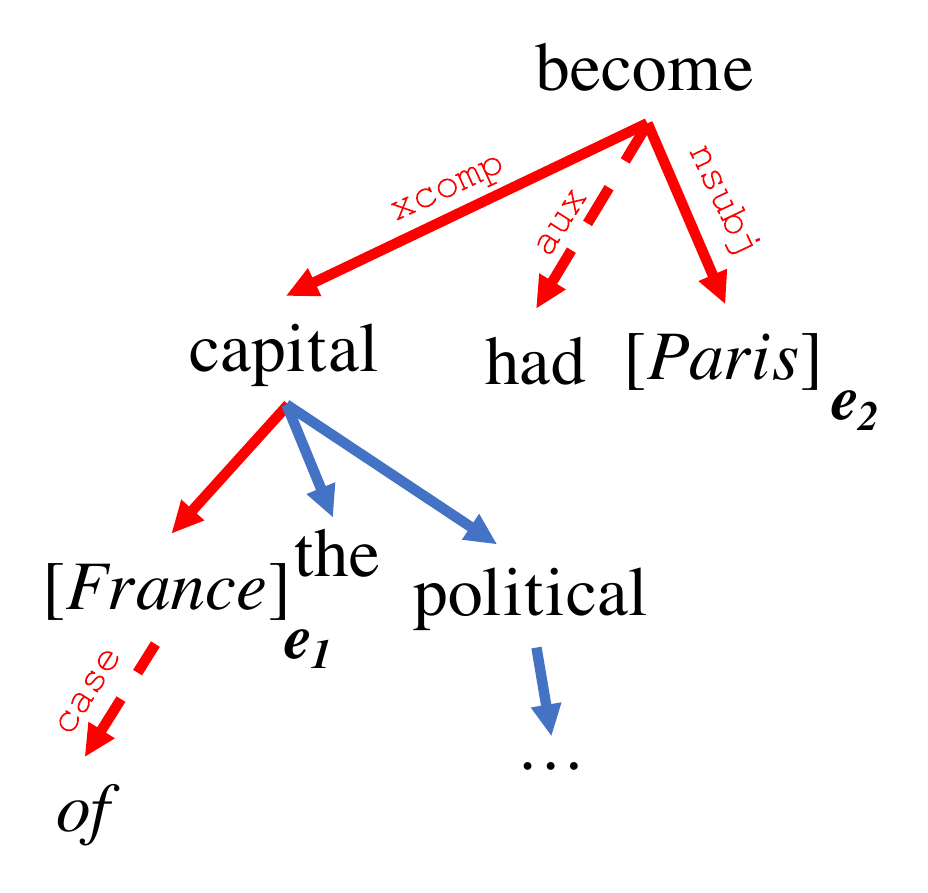}
	\caption{Finding semantic path}
	\label{fig:dep_tree1}
	\end{subfigure}
	\begin{subfigure}[t]{0.49\linewidth}
	\centering
	\includegraphics[width=0.8\linewidth]{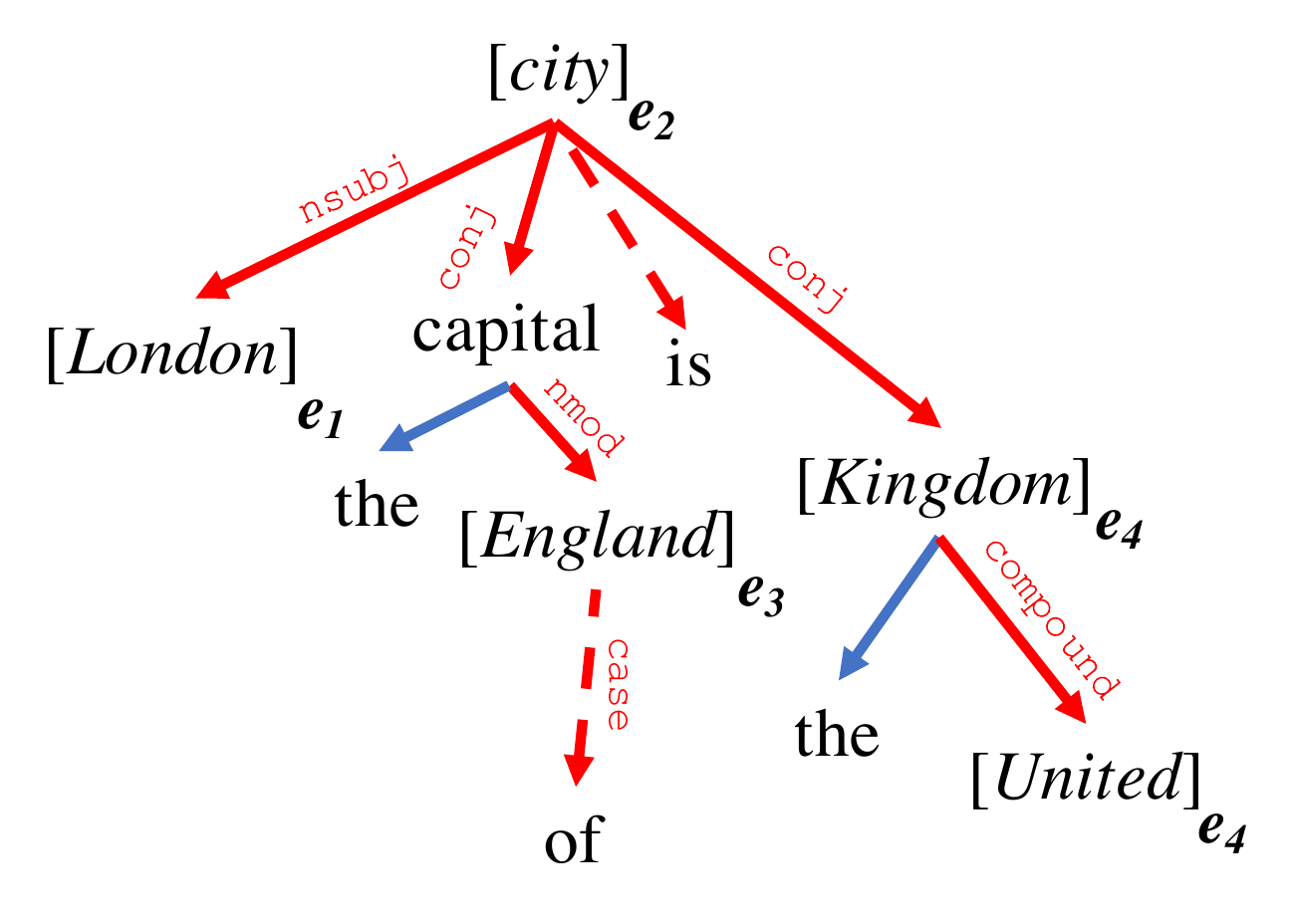}
	\caption{Positive Entity Pairs Initialization}
	\label{fig:dep_tree2}
    \end{subfigure}
\vspace{-1em}
  \caption{Dependency parsing tree of example sentences $s_1$ and $s_3$ in Fig.~\ref{fig:Framework}, Segmented entities are marked as ``[entity\_token]$_{e_i}$''}
  \label{fig:dep_tree}
\end{figure}

%To reduce noise from irrelevant semantic phrases or clauses in long sentences with multiple entities, leveraging information along the dependency path between two given entities has been proved useful for open information extraction~\cite{Xu2013OpenIE,Gamallo2012DependencyBasedOI}.
\xhdr{Generating Candidate Entity Pairs.}
\label{sec:positive-pairs}
For a given sentence s, after phrase segmentation, we have entity phrases $e_1, e_2, ..., e_n$ and relation phrases $r_1, r_2, ..., r_n$. However, it's computationally intractable to explore possible relationships between every entity pair and a large portion of tuples are incorrect among $n(n-1)$ pairs. $E_p^+$ are candidate entity phrase pairs. Here we heuristically initialize ${E_p^+}^0$ by attaching the \textit{nearest} subject $e_i$(within the sentence) to the object $e_j$ and make an approximation that each entity argument phrase can only be an object once; this also guarantees entity pairs to be distinctive.
The nearest subject of $e_j$ is defined as the entity $e_i$ that has the shortest dependency path length to $e_j$ among all other entities. Considering Fig.~\ref{fig:dep_tree2}, we would like to find the subject of entity $e_4:$ \textit{United Kingdom}, the lengths of the shortest paths between $e_4$ and $e_1$, $e_2$, $e_3$ are 2,3,1.
For those entity candidates with the same distance, see Fig.~\ref{fig:dep_tree2}, both $e_1$: \textit{London} and $e_4$: \textit{United Kingdom} is one hop away from $e_2$: \textit{city}. In this situation, we will prefer the subject with ``nsubj'' type \ie $e_1$. If there are still multiple entities, we will choose the closest entity in the original sentence.

\xhdr{Semantic Path Generation.}
Once ($e_h$, $e_t$) $\in$ $E_p^+$ is determined, the semantic path is defined as the shortest dependency path between two arguments. Compared with using word sequence between ($e_h$, $e_t$) directly, the semantic path helps cloud irrelevant information. For example, in Fig.~\ref{fig:dep_tree1}, the semantic path between \textit{``Paris''} and \textit{``France''} of sentence $s_3$ is marked in red, where word sequence``the political, economic...'' is correctly excluded. To preserve integrity of potential relation phrases, we further include particles and preposition along the shortest dependency path as part of the semantic path, which is shown as red dotted line in Fig.~\ref{fig:dep_tree1}.

\begin{definition}
\textbf{(Semantic Path)}
\textsl{For an entity phrase pair ($e_h$, $e_t$) in the same sentence, the semantic path is defined as word sequence \textit{$SP_{h,t}$} along expanded dependency path.}
\end{definition}

\begin{footnotesize}
\begin{example}[Generating Relation Tuples] Extracting relation phrases on the semantic path
\label{eg:tuple_generation}
\begin{center}
\textbf{[Paris]} + \underline{had become} + \underline{capital of} + \textbf{[France]}
\end{center}
\end{example}
\end{footnotesize}

%Instead of words, we treat phrases as the smallest semantic unit, and each predicate $p_{h,t}$ is composed of one or more relation phrases $r$.
We now present how we generate valuable relation tuples according to semantic path, \ie
\begin{equation} \label{eq:tuple_generation}
\begin{split}
\P(r, e_h, e_t) = \prod_{i=1}^{n} \P(r_i|s_i, e_h, e_t) \P(s_i|b_i,b_{i+1}) \\
\max_{p_{h,t}} \P(r, e_h, e_t) \Rightarrow \Sum_{r_i \in p_{h,t}} \log \sigma(r_i, e_h, e_t) + \log w_i
\end{split}
\end{equation}
where $b_1,b_2,...,b_{n+1}$ are boundary index along semantic path $SP_{h,t}$ of entity phrase pair$(e_h, e_t)$. $\P(s_i|b_i,b_i+1)$ is inherited from phrase extraction module as sentence segmentation probability $w_i$, then \ReMine judges whether it is a good relation between entity $e_h$ and entity $e_t$. 
%In the sentence $s_4$, we extract relation tuple $\langle$ ``Your dry cleaner'', ``\textbf{set out, on}'', ``foot'' $\rangle$, since both relation phrases \textbf{set out, on} are coherent with global measure $\sigma$ and $w_i$. 
Notice that the relation phrase boundary $i \in p_{h,t}$ in equation~\ref{eq:tuple_generation} can be derived via dynamic programming since $w_i$ and $\sigma$ is known for every possible segmentation. In example~\ref{eg:tuple_generation}, within entity pair $\langle$\textit{Paris}, \textit{France}$\rangle$, the semantic path is ``\textsf{had become capital of}''. Tuple $\langle$\textit{Paris}, \textsf{had become$|$capital of}, \textit{France}$\rangle$ will be generated as both relation phrases \textsf{had become} and \textsf{capital of} are coherent with global cohesiveness measure $\sigma$ and $w_i$.

%\subsection{Global Measuring of Structural Cohesiveness}
\subsection{Global Cohesiveness Module}
\label{sec:global_measure}
Inevitably, false tuple like $\langle$\textit{city}, \textsf{is capital of}, \textit{England}$\rangle$ will be generated by relation tuple generation module as introduced in Sec~\ref{sec:tuple_generation}, since the nearest subject of \textit{England} in Fig.~\ref{fig:dep_tree2} is \textit{city}. To get rid of such false tuples, current methods use textual patterns~\cite{schmitz2012open,del2013clausie} to identify it as a false extraction. In contrast, we design global cohesiveness measure using corpus-level statistics, and integrate the measure with the relation tuple generation. To capture the global cohesiveness of relation tuples, we adopt translation-based multi-relational data representation~\cite{bordes2013translating}.
\begin{equation} \label{eq:global-measure}
\sigma(p, h, t) = - d(h+p, t) ; \quad d(h+p, t) =  \norm{v_h+v_p-v_t}
\end{equation} 
where $v_h,v_t$ are embeddings for head and tail entities, $p$ is the predicate. Such a measure imposes reliable relation tuples on small translating distance between $h+p$ and $t$. We use $L_1$ norm in \ReMine.

Based on initial positive entity pairs constructed ${E_p^+}^0$ and relation tuples $\T$, we construct a pseudo knowledge graph. Particularly, predicate $p_{h,t}$ may contain several relation phrases. Motivated by process of knowledge traverse~\cite{guu2015traversing}, we average multiple relation phrases embeddings to represent the predicate \ie $v_p = \sum_{i=1}^n v_{r_i}/n$.

\begin{footnotesize}
\begin{example}[Generating False Tuples]
\label{eg:false_generation}
\begin{center}
$\langle$\textit{Paris}, \textsf{become capital of}, \textit{France}$\rangle$ $\rightarrow$ $\langle$ \textbf{\textit{city}}, \textsf{become capital of}, \textit{France}$\rangle$, $\langle$\textit{Paris}, \textsf{become capital of}, \textbf{\textit{Burger King}}$\rangle$
\end{center}
\end{example}
\end{footnotesize}

In order to learn a global cohesiveness representation $\mathcal{V}$, we construct correlated negative tuples from positive seeds. For instance, as seen for example~\ref{eg:false_generation}, we see that for a positive tuple, we can generate many incorrect or ``negative'' tuples.

The cohesiveness measure $\sigma$ is optimized by maximizing the cohesiveness margin between positive and negative tuples,
\begin{equation}
\begin{split}
\max_{\mathcal{V}} \Sum_{p,h,t}^\T \Sum_{p,h',t'}^{\T^-} [\sigma(p,h,t)- \sigma(p,h',t')-\gamma]_{-}
\end{split}
\end{equation}
where $[x]_{-}$ denotes the negative part of x, $\T$ denotes positive relation tuples generated by local relational extraction, $\gamma$ is the hyper margin, $(p,h',t') \in \T^-$ is composed of training tuples with either $\h$ or $\t$ replaced.

\subsection{The Joint Optimization Problem}
Relation tuple generation in Section~\ref{sec:tuple_generation} incorporates cohesiveness similarity $\sigma$. Additionally global cohesiveness measure learning depends on extracted tuples $\T$. We now show how local context and global cohesiveness introduced above can be integrated.

\xhdr{Overall Updating Schema.} The final objective for update is formulated as the sum of both sub-objectives,
\begin{eqnarray}
&\max_{\mathcal{V},\T} \mathcal{O} = \mathcal{O}_{local} + \mathcal{O}_{global} \\
&\mathcal{O}_{local} = \Sum_{E_p^{+}} \log \P(p_{h,t}, e_h, e_t)\label{eqn:local_obj} \\
&\mathcal{O}_{global} = \Sum_{p,h,t}^\T \Sum_{p,h',t'}^{\T^-} [\sigma(p,h,t)- \sigma(p,h',t')-\gamma]_{-}\label{eqn:global_obj}
\end{eqnarray}
To maximize the above unified open IE objective, see Alg.~\ref{algorithm:ReMine}, we first initialize positive entity pairs ${E_p^+}^0$. Given entity phrase pairs, we perform local optimization, which leads to positive relation tuples $\T$.
Note that, at the first round, there is no global representation, so we initialize all $\sigma=1$ as identical. Then we update global phrase semantic representation via stochastic gradient descent. With both global cohesiveness information and local segmentation results, \ReMine updates relation tuples as described in Alg.~\ref{algorithm:ReMine}. 
Overall \ReMine solves the integrated problem greedil and it iteratively updates local and global objectives until a stable $E_p^+$ is reached.

%Having global measuring~\ref{eq:global-measure}, we can capture the semantic drift in one predicate. Sometimes, predicate may be composed multiple relation phrases, whereas some of them are not crucial for the relation tuple \eg ``and''. Instead manually constructing pruning rules, we solve the problem by finding important relation phrases in predicate.
%\begin{eqnarray}
%\argmin_{\r \subset \R}\norm{v_h-v_t+v_\r} \\
%\D(\R | e_i, e_j, P_{(e_i,e_j)}) = \min \norm{v_{e_i} + v_\r - v_{e_j}}
%\end{eqnarray}
\begin{footnotesize}
\begin{example}[Updating Relation Tuples] In sentence $s_1$,\hfill
\label{eg:update_tuples}
\begin{center}

$\langle$\textit{city}, \textsf{is capital of}, \textit{England}$\rangle \rightarrow \langle$\textit{London}, \textsf{is capital of}, \textit{England}$\rangle$
\end{center}
\end{example}
\end{footnotesize}
\xhdr{Update Positive Entity Pairs and Relation Tuples.}
\label{sec:update_tuples}
Given a semantic representation for each entity $\e$ and relation $\r$ and local segmentation between entity pairs, we can update the \textit{Positive Entity Pairs} by finding the  most semantically consistent subject $e_h$ for each object $\e_t$ among $M_{sp}$-nearest neighbors on the dependency parsing tree. By optimizing $\P(r, e_h, e_t)$ in Eq.~\ref{eq:tuple_generation}, we also obtain the relation tuples for updated positive pairs ${E_p^+}^{n+1}$.   
\begin{equation}
E_p^{+} = \argmax_{e_h} \P(p_{h,t}, e_h, e_t)
\end{equation}
%In corpus, \textit{London} and \textit{Paris} share lots of predicate and tail entities. 
In example~\ref{eg:update_tuples} and Fig.~\ref{fig:dep_tree2}, false tuple $\langle$\textit{city}, \textsf{is capital of}, \textit{England}$\rangle$ will be updated as $\langle$\textit{London}, \textsf{is capital of}, \textit{England}$\rangle$. Seeing \textit{London} and \textit{Paris} share lots of predicate and tail entities, the updated tuple is more cohesive with others \eg $\langle$\textit{Paris}, \textsf{become capital of}, \textit{France}$\rangle$. 

\begin{algorithm}[t]
\begin{scriptsize}
    	\KwIn{corpus $\D$, sentence {S}, text features {$\F$}, convergence threshold $t$} 
		\KwOut{relation tuples $\T$, semantic representation $\mathcal{V}$, similarity measure $\sigma$}
		generate entity and relation seeds via distant corpus linking \;
		\textbf{phrase extraction module} outputs entity phrases, relation phrases, sentence segmentation probability $\mathcal{W}$ \;
		initialize positive $E_p^{+0}$, cohesiveness measure $\sigma= 1$ \;
		generate relation tuples $\T$ among  $E_p^{+0}$ \;
		\Do{$\frac{\Delta_E}{|E_p^{+n}|} > t$}{
		    update $\mathcal{V},\sigma$ in Eq.~(\ref{eqn:global_obj}) via \textbf{global cohesiveness module} \;
		    $E_p^{+n} \leftarrow \emptyset$, $\Delta_E \leftarrow 0$ \;
			\For{each tuple $\langle e_h, p_{ht}, e_t \rangle \in \T$} {
			    construct candidate subject sets \textbf{s} of $e_t$ with at most $M_{sp}$ entities\;
			    $\sigma_* \leftarrow \sigma(e_h, p_{h,t}, e_t)$ \;
			    \For{i = 1 to $M_{sp}$}{
				generate $\langle s_i, p_{i,t}, e_t \rangle$ in Eq.~(\ref{eqn:local_obj}) via \textbf{relation tuple generation module} given $\mathcal{W}$ and $\mathcal{V}$\;
				\If{$\sigma(s_i, p_{i,t}, e_t) > \sigma_*$}{
				    $\sigma_* \leftarrow \sigma(s_i, p_{i,t}, e_t)$, $e_*\leftarrow s_i$ \;
				    }
				}
				$E_p^{+n} \leftarrow E_p^{+n} \bigcup \langle e_*, e_t \rangle$\;
				\If{$e_* \neq e_h$}{
				    $\Delta_E \leftarrow \Delta_E + 1$, $\langle e_*, p_{*t}, e_t \rangle \leftarrow \langle e_h, p_{ht}, e_t \rangle$  update $\T$ \;
				}
			}
		}
		%\doWhile{$E_p^+$ is not stable}
        \caption{The \ReMine Algorithm for Joint Optimization}
        \label{algorithm:ReMine}
\end{scriptsize}
\end{algorithm}

%\xhdr{Overall Updating Schema.} From an overall point of view, the final objective for update is formulated as the sum of both sub-objectives:
%\begin{equation}
%\mathcal{O} = \mathcal{O}_{local} + \mathcal{O}_{global}
%\end{equation}
%To maximize above unified open IE objective, see Alg.~\ref{algorithm:ReMine}, we first initialize positive entity pairs ${E_p^+}^0$. Given entity argument pairs, we perform local optimization, which leads to positive relation tuples $E_{h,l,t}^+$.
%Note that, at the first round, there is no global representation, so we initialize all $w_i=1$. Then we update global phrase semantic representation via stochastic gradient descent. With both global cohesiveness information and local segmentation result, \ReMine updates positive pairs as described in Sec.~\ref{sec:positive-pairs}. \ReMine solves the integrated problem in a greedy manner, first fix $\mathcal{W}$, Relation Tuple Generation Module selects positive $\mathcal{T}$. Then we maximize global objective by updating $\mathcal{W}$ and $\mathcal{V}$. Finally, fix $\mathcal{W}$, update positive pairs is identical to select new set of $\mathcal{A}$. We iteratively updating local and global objectives until the convergence, which will lead to a stable $\mathcal{W}$, $\mathcal{V}$ and $E_p^+$.
%!TEX root = main.tex
\begin{table}[t]
	\caption{Features used in the phrase extraction module (Sec.~\ref{sec:local_seg}).}
	\vspace{-0.3cm}
	\label{tab:features}
	\small
	\begin{tabular}{|l|l|}
		\hline
		\textbf{Feature} & \textbf{Descriptions} \\
		\hline
		popularity & raw frequency, occurrence probability  \\
		
		completeness & whether can be interpreted as a complete semantic unit \\
		
		concordance & tokens in quality phrases should co-occurs frequently \\
		
		punctuation & phrase in parenthesis, quote or has dash after \\
		
		stopwords & first/last token is stopword and stopword ratio \\
		
		word shape & first capitalized or all capitalized \\
		
		POS tags & unigram and bigram POS tags \\
		\hline
	\end{tabular}
\end{table} 
%\vspace{0.2cm}

\section{Experiments}
For thorough evaluation of the proposed approach, we test the performance of \ReMine system from two aspects, \ie, quality of the extracted entity phrases (\ie, entity phrase extraction with distant training), and quality of the extracted relation tuples (\ie, output of the Open IE system). 
%We compare the output of our system with state-of-the-art entity phrase extraction methods and Open IE systems. 
In particular, we compare \ReMine with state-of-the-art Open IE systems to validate our three claims: (1) the domain-independent framework performs consistently well on different domains, (2) global structure cohesiveness improves performance of Open IE, and (3) the proposed iterative updating algorithm is effective and scalable.
%To examine our claim: (1) a domain-specific and end-to-end mannered pipeline performs consistently well on different domains; and (2) global structure cohesiveness improves Open Information Extraction. (3) Iterative updating local and global objective is scalable and effective.
%Clear entity boundaries and high-quality extractions can lead to successful downstream applications including knowledge base construction and question answering.
%In terms of weakly-supervised entity phrase extraction, we provide seed entities using distant supervision in conjunction with knowledge bases. Since entity seeds in various type like \textsl{architecture, tv, business, interests, award, book, \etc}, we expect to show phrase extraction module are robust to noise and can provide accurate entity phrase extraction result. For relation tuple extraction, we expect relation tuple generation module can generate reliable and diverse facts.

\subsection{Experimental Setup}
\xhdr{Datasets.}
We use three public datasets\footnote{Codes and datasets can be downloaded at \url{https://github.com/GentleZhu/ReMine}} from different domains in our experiments: (1) NYT~\cite{riedel2013relation}: a corpus consisting of 23.6k sentences from $\sim$294k 1987-2007 New York Times news articles. 395 sentences are manually annotated with entities and their relationships. 
(2) Wiki-KBP~\cite{ling2012fine}: The training corpus contains 2.4k sentences sampled from $\sim$780k Wikipedia articles~\cite{ling2012fine} as the training corpus and 290 manually annotated sentences as test data.
(3) Twitter~\cite{zhang2016geoburst}: consists of 1.4 million tweets from Los Angeles with entities and/or noun phrases collected from 2014.08.01 to 2014.11.30. 

\xhdr{Distant Supervision for Generating Training Data.}
\label{sec:distant-linking}
For each corpus, we apply the entity linking tool DBpedia Spotlight \footnote{\url{https://github.com/dbpedia-spotlight/dbpedia-spotlight}}~\cite{isem2013daiber} to recognize DBpedia entities in sentences and use them as ``seed" entity phrases. With seed entity phrases, we generate relation phrases between each pair of entity mentions via pattern matching (see Sec.~\ref{sec:tuple_generation}), forming the seed relation tuples. These seed tuples are used as distant supervision for training segmentation algorithm (thus ``distant training"). We then follow the procedure introduced in Sec.~\ref{sec:local_seg} to segment sentences into entity and relation phrases.

\xhdr{Phrase Features Generation.}
%add random forest classifiers here
In order to estimate quality and catgeory of phrases, we use features $\mathcal{F}$ in Table~\ref{tab:features}. These features can be grouped into several different categories, \ie statistic features, token-wise features and POS features. \ReMine treats phrases with multiple POS tag sequences as different patterns. For example, ``work NN'' and ``work VBP'' are two different semantic patterns. 
%Shang et al. ~\cite{shang2017automated} show that considering POS tags in quality predictor yields better performance.
%Compared with previous phrase mining work, we introduce extra linguistic constraints as prior of phrase segment to guide phrasal segmentation - dependency parsing tree patterns, which usually bring rich context information.
%For example, for the sentence ``Gov. Tim Pawlenty of Minnesota order the ...'', \ReMine would segment ``[Gov. Tim Pawlenty of Minnesota]'' together as a whole entity phrase rather than ``[Gov. Tim Pawlenty] [of Minnesota]'' since the context-dependent prior prefers one complete tree pattern rather than two separate ones.
We applied the Stanford CoreNLP~\cite{manning2014stanford} tool to get POS tags and dependency parsing trees. 
%We use the same external linguistic features as other Open IE methods in our experiments.

\xhdr{Compared Methods.} 
For the entity phrase extraction task, NYT and Wiki-KBP are used for evaluation, since both datasets contain annotated entity mentions in test set.
We adopt the sequence labeling evaluation setup~\cite{ma2016end}, and compare \ReMine's entity phrase extraction module with two state-of-the-art sequence labeling methods and one distantly-supervised phrase mining method on the test sets: (1) \textbf{Ma \& Hovy}~\cite{ma2016end}: adopts a Bi-directional LSTM-CNN structure to encode character embeddings and pre-trained word embeddings; (2) \textbf{Liu.~\etal}~ \cite{2017arXiv170904109L}: incorporates a neural language model and conducts multi-task learning to guide sequence labeling; and (3) \textbf{AutoPhrase}~\cite{shang2017automated}: the state-of-the-art quality phrase mining method with POS-guided phrasal segmentation. 
%We will not compare \ReMine with existing Open IE algorithms on entity phrase extraction, since they do not explicitly output entity phrases and same entity will have various boundaries in one sentence.

For the relation tuple extraction task, we consider following Open IE baselines for comparison: (1) \textbf{OLLIE}~\cite{schmitz2012open} utilizes open pattern learning and extracts patterns over the dependency path and part-of-speech tags. (2) \textbf{ClausIE}~\cite{del2013clausie} adopts clause patterns to handle long-distance relationships. (3) \textbf{Stanford OpenIE}~\cite{angeli2015leveraging} learns a clause splitter via distant training data. (4) \textbf{MinIE}~\cite{gashteovski2017minie} refines tuple extracted by ClausIE by identifying and removing parts that are considered overly specific. (5) {\bf ReMine-L} is a base model of our approach with only local context. We only plot precision@300 in Fig.~\ref{fig:pr} as no ranking measure is deployed.
(6) {\bf ReMine-G} extend \ReMineL by ranking tuples via global cohesiveness without updating entity phrase pairs and any further iterations. (7) {\bf ReMine} is our proposed approach, in which relation tuple generation module collaborates with global cohesiveness module.

\xhdr{Parameters Settings.} For baselines of entity phrase extraction task, we tune all the models using the same validation set. 
In the testing of \ReMine and its variants, we set hypermargin $\gamma$ = 1, maximal phrase length $\epsilon = 6$, number of candidate subject entity phrase for each tail entity $M_{sp}=6$ and learning rate of the global cohesiveness module $\alpha = 10^{-3}$. The dimension of global cohesiveness representation k is 100. We stop further joint optimization if the ratio $\t$  of updated tuples is smaller than $10^{-3}$.

%Since \ReMine approaches information extraction task in two steps, \ie phrase extraction (or entity phrase extraction) and relation tuple extraction, we mainly compare \ReMine with other baselines for these two tasks.

%

%\begin{table}[th]
%	\centering
%	\caption{Extra supervision required by different systems.}
%	\label{tab:fea}
%	\begin{tabular}{|m{1.3cm}|c|c|c|c|}
%		\hline
%		Compared & Lexical & Dependency & Distant  & Human  \\
%		Methods         & Patterns         & Parsing            & Supervision         &  %Curation      \\ \hline
%		ClausIE         &  No              & Yes                & No                  & %No             \\ \hline
%		Stanford OpenIE & Yes              & Yes                & Yes                 & %No             \\ \hline
%		OLLIE           & Yes              & Yes                & No                  & %Yes            \\ \hline
%		ReMine          & No               & Yes                & Yes                 & No             \\ \hline
%	\end{tabular}
%\end{table}
 
\begin{table}[ht!]
\begin{small}
	\centering
	\caption{Performance comparison with state-of-the-art entity phrase extraction algorithms for the weakly-supervised entity phrase extraction task.}
	\vspace{-0.3cm}
	\label{tab:entity-detection}
	\begin{tabular}{|r||c|c|c||c|c|c|}
		\hline
		\multirow{2}{*}{\textbf{Methods}}& \multicolumn{3}{c||}{\textbf{NYT} \cite{riedel2013relation}} & \multicolumn{3}{c|}{\textbf{Wiki-KBP} \cite{ling2012fine}} \\ \cline{2-7}
		& F1 & Prec & Rec & F1 & Prec & Rec\\
		\cline{2-7}
		\hline
		AutoPhrase \cite{shang2017automated} & 0.531 & 0.543 & 0.519 & 0.416 & 0.529 & 0.343\\
		\hline
		Ma \& Hovy \cite{ma2016end} & 0.664 & 0.704 & 0.629 & 0.324 & 0.629 & 0.218  \\
		\hline
		Liu. \etal \cite{2017arXiv170904109L} & \textbf{0.676} & \textbf{0.704} & 0.650 & 0.337 & 0.629 & 0.230\\
		\hline
		\ReMine & 0.648 & 0.524 & \textbf{0.849} & \textbf{0.515} & \textbf{0.636} & \textbf{0.432} \\
		\hline
	\end{tabular}
	\end{small}
\end{table}

\xhdr{Cut-off Threshold for Extraction Output.} The number of tuple extractions from different systems can vary a lot. For example, for the first 100 sentences in the NYT test set, both \ReMine and OLLIE get about 300 tuples. In contrast, Stanford OpenIE returns more than 1,000 tuples. However, many paraphrased extractions can be found within them.
Since each extracted tuple is also assigned with a confidence score, we select 300 tuples for both datasets with the highest scores for each open IE system to report the performance. By selecting 100 sentences from NYT test set and 300 tweets from Twitter test set, we believe $\sim$3 tuples per sentence in News domain and $\sim$1 tuple per sentence in Twitter are reasonable for a fair comparison. A more detailed study can be found in Sec.~\ref{sec:analysis_distinct}.

\xhdr{Annotation of Ground-truth Data.} We manually labeled the top-300 tuple extraction results obtained from all compared methods via pooling method (\ie, high-confidence tuples by each system are pooled together as the candidate set). Each extracted tuple in the candidate set was labeled by two independent annotators. A tuple is labeled as positive only if both labelers agree on its correctness. All tuples with conflicting labels results were filtered. Similar to~\cite{del2013clausie}, we ignored the context of the extracted tuples during labeling. For example, both (\textit{``we", ``hate", ``it"}) and (\textit{``he", ``has", ``father"}) will be treated as correct as long as they meet the fact described in the sentence. However, tuples cannot be read smoothly will be labeled as incorrect propositions. For example, (\textit{``he", ``is", ``is the professor"}) and (\textit{``he", ``is", ``the professor and"}) will not be counted since they have mistakes at the word segmentation level. The Cohen's Kappa value  between the two labelers are 0.79 and 0.73 for the NYT dataset and the Twitter dataset respectively.
%\xhdr{Evaluation Setup.}
%Here we describe the setups for two tasks.

%\xhdr{(1) Weakly-supervised entity phrase extraction.} For this task, NYT and Wiki-KBP are used for evaluation, since both datasets contain manually-annotated entity mentions in test set. The training data is generated through distant supervision described above without type information. Considering distant supervision may not be as a gold-standard annotation, we use high-confidence seeds for \ReMine and other baselines, pushing them towards a fair comparison.
%We use Precision (\ie how many entities we get are correct), Recall (\ie how many correct entities do we get), and F1-score to evaluate the performances.
\xhdr{Evaluation Metrics.}
We use Precision (\ie how many entities we get are correct), Recall (\ie how many correct entities do we get), and F1-score to evaluate the performances on entity phrase extraction task, same as other sequence labeling studies~\cite{ma2016end}.
%\xhdr{(2) Relation tuple extraction.}
%We aim to compare performance on both normal and short text, so we choose NYT and Twitter dataset for evaluation. 
For the Open IE task, since each tuple obtained by \ReMine and other benchmark methods will also be assigned a confidence score. We rank all the tuples according to their confidence scores. Based on the ranking list, we use the following four measures: $P@k$ is the precision at rank $k$. $MAP$ is mean average precision of the whole ranking list. $NDCG@k$ is the normalized discounted cumulative gain at rank k. $MRR$ is the mean reciprocal rank of the whole ranking list. 
Note that we do not use recall in this task because it is impossible to collect all the ``true" tuples.

%Fixed phrase boundaries prevent \ReMine from generating redundant facts from the corpus, a detailed study can be found in Section~\ref{sec:analysis_distinct}.

\begin{table*}[th]
	\centering
	\caption{Performance comparison with state-of-the-art Open IE systems on two datasets from different domains, using Precision@K, Mean Average Precision (MAP), Normalized Discounted Cumulative Gain (NDCG) and Mean Reciprocal Rank (MRR).}
	\vspace{-0.3cm}
	\label{tab:performance}
	\begin{small}
		\begin{tabular}{|r||c|c|c|c|c|c||c|c|c|c|c|c|}
			\hline
			\multirow{2}{*}{\textbf{Methods}}& \multicolumn{6}{c||}{\textbf{NYT} \cite{riedel2013relation}} & \multicolumn{6}{c|}{\textbf{Twitter} \cite{zhang2016geoburst}} \\ \cline{2-13}
			\       & P@100 & P@200 & MAP & NDCG@100 & NDCG@200 & MRR & P@100 & P@200 & MAP & NDCG@100 & NDCG@200 & MRR \\
			\hline
			ClausIE & 0.580 & 0.625 & 0.623 & 0.575 & 0.667 & 0.019 & 0.300 & 0.305 & 0.308 & 0.332 & 0.545 & 0.021 \\\hline
			Stanford & 0.680 & 0.625 & 0.665 & 0.689 & 0.654 & 0.023 & 0.390 & 0.410 & 0.415 & 0.413 & 0.557 & 0.023 \\\hline
			OLLIE  & 0.670 & 0.640 & 0.683 & 0.684 & 0.775 & \textbf{0.028} & 0.580 & 0.510 & 0.525 & 0.519 & 0.626 & 0.017 \\\hline
			MinIE & 0.680 & 0.645 & 0.687 & 0.724 & 0.723 & 0.027 & 0.350 & 0.340 & 0.361 & 0.362 & 0.541 & \textbf{0.025} \\\hline
			\ReMineG  & 0.730 & 0.695 & 0.734 & 0.751 & 0.783 & 0.027 & 0.510 & 0.580 & 0.561 & 0.522 & 0.610 & 0.021 \\\hline
			\ReMine  & \textbf{0.780} & \textbf{0.720} & \textbf{0.760} & \textbf{0.787} & \textbf{0.791} & 0.027 & \textbf{0.610} & \textbf{0.610} & \textbf{0.627} & \textbf{0.615} & \textbf{0.651} & 0.022 \\
			\hline
		\end{tabular}
		\end{small}
		\vspace{0.3cm}
\end{table*}

%!TEX root = main.tex
\subsection{Experiments and Performance Study}
\xhdr{1. Performance on Entity Phrase Extraction.}
\label{exp:ner}
The training data is generated through distant supervision described above without type information.
%Although pre-trained models~\cite{ma2016end,2017arXiv170904109L} yield extremely high precision in standard sequence labeling tasks \eg NER, POS Tagging, NP Chunking. 
Regarding open domain extractions, we train baseline models using the same distant supervision as \ReMine, to push them towards a fair comparison. Table~\ref{tab:entity-detection} demonstrates the comparison result over all datasets. In the Wiki-KBP dataset, \ReMine evidently outperforms all the other baselines. In the NYT dataset, \ReMine has a rather high recall and is on par with the two neural network models on F1-score.

\begin{figure}[t!]
	\centering
	\begin{subfigure}[t]{0.8\linewidth}
		\includegraphics[width=1.01\linewidth]{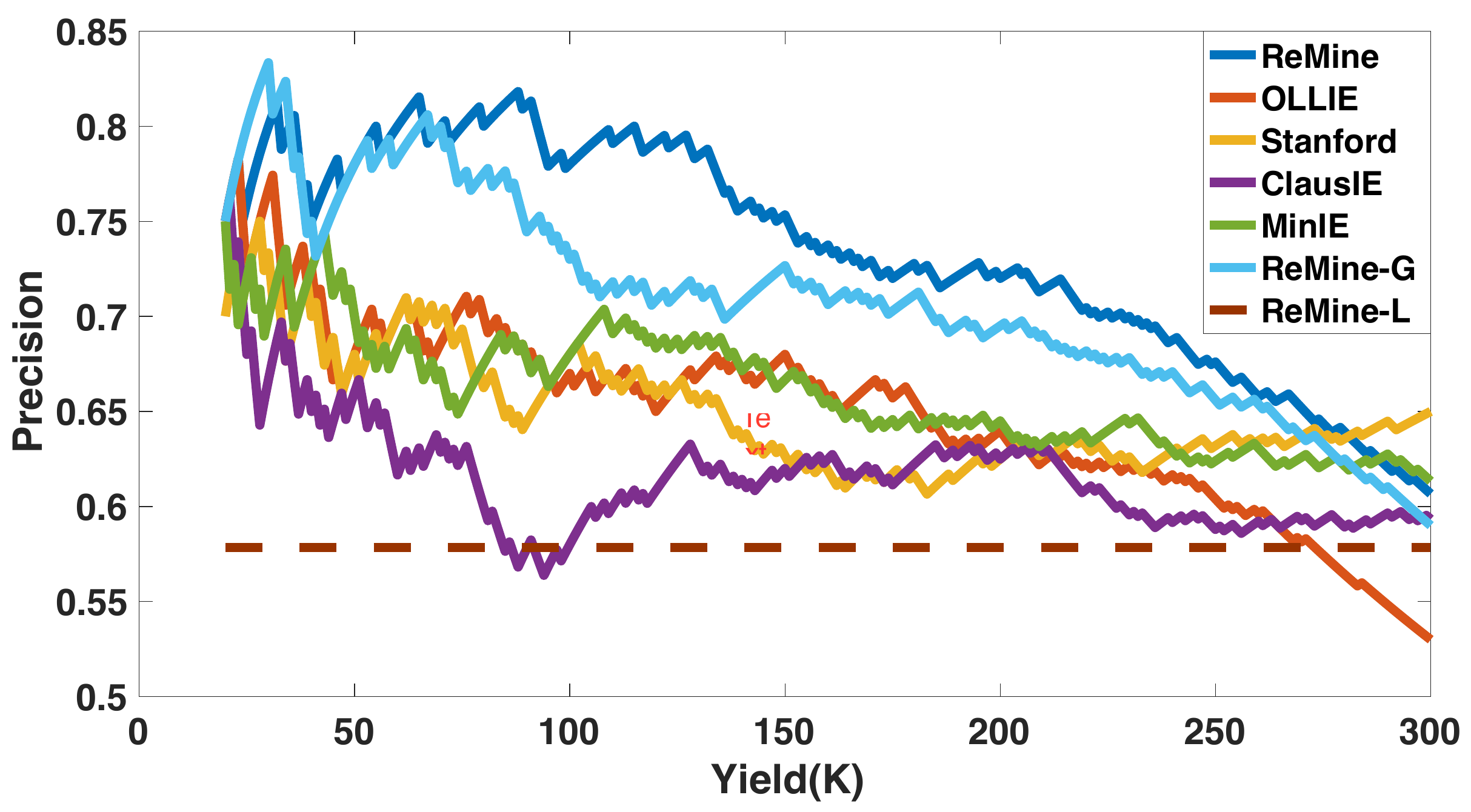}
		\vspace{-0.6cm}
		\caption{NYT}
		\label{fig:nyt_pr}
	\end{subfigure}
	\begin{subfigure}[t]{0.8\linewidth}
		\includegraphics[width=1.01\linewidth]{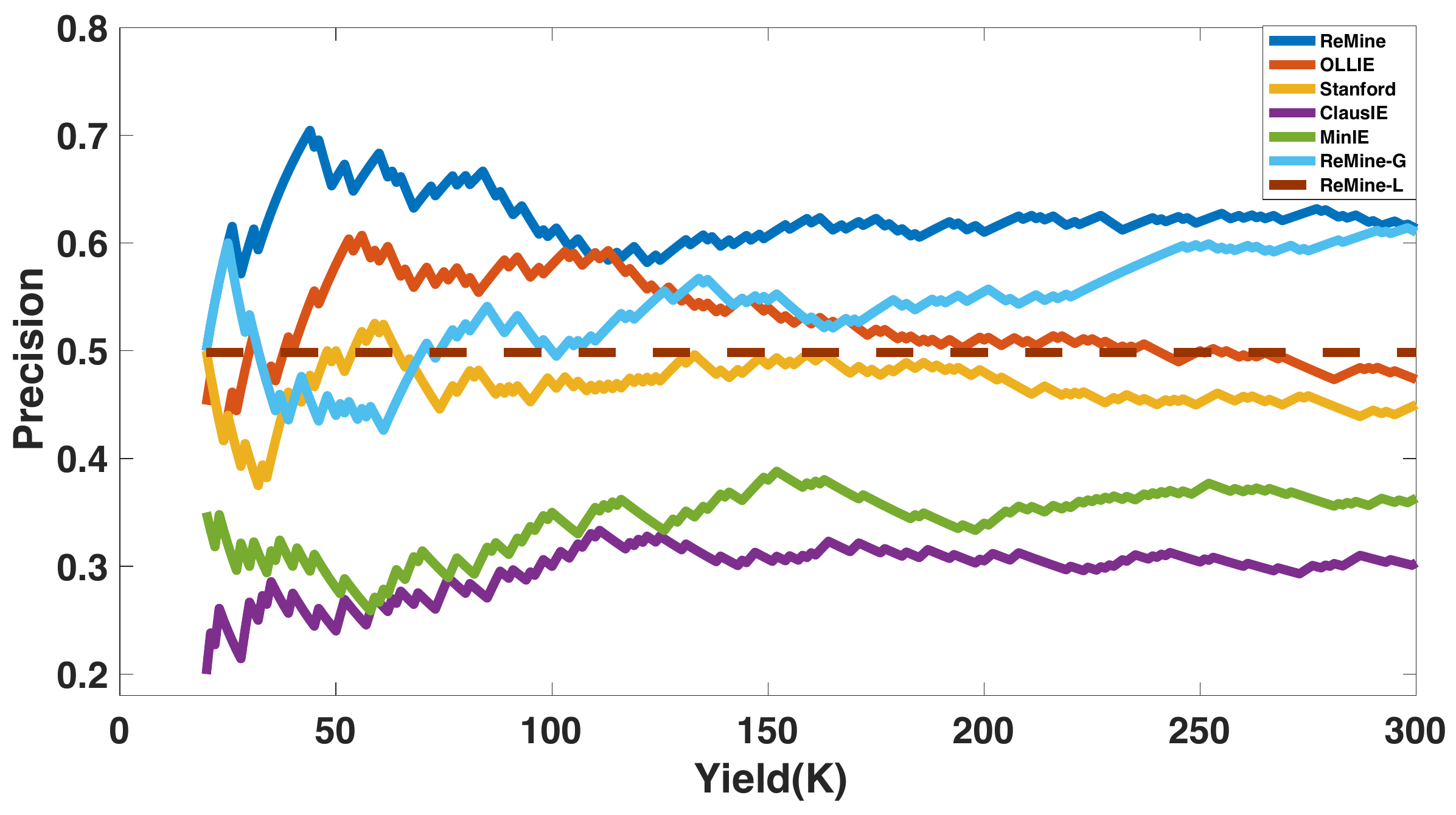}
		\vspace{-0.6cm}
		\caption{Twitter}
		\label{fig:twitter_pr}
	\end{subfigure}
	\vspace{-0.2cm}
	\caption{The Precision@K curves of different open IE systems on NYT and Twitter datasets.}
	\label{fig:pr}
	%\vspace{-0.5cm}
\end{figure}

\xhdr{2. Performance on Relation Tuple Extraction.} 
On NYT and twitter test set, we compare \ReMine with its variants \ReMineL and \ReMineG as well as four baseline open IE systems mentioned above. The results are shown in Figure~\ref{fig:pr} and Table~\ref{tab:performance}.
%The scores were shown in Table~\ref{tab:kappa}.

%Among all the open IE system described above, \ReMine and OLLIE extract a relatively small number of tuples. For example, for the first 100 sentences in the NYT test set, both \ReMine and OLLIE get about 300 tuples. In contrast, Stanford OpenIE returns more than 1,000 tuples. To alleviate the ``unintentional paraphrasing'' issue, since each extracted tuple is also assigned a confidence score, we select 300 tuples for both datasets with the highest scores for each open IE system to plot the curves. By selecting 100 sentences from NYT test set and 300 tweets from Twitter test set, we believe $\sim$3 tuples per sentence in News domain and $\sim$1 tuple per sentence in Twitter seems to be reasonable. 

\medskip
``\textit{Does \ReMine perform consistently well on different domains?}''
\smallskip

According to the curves in Figure~\ref{fig:nyt_pr} and~\ref{fig:twitter_pr}, \ReMine achieves the best performance among all open IE systems. All methods experience performance drop in Twitter, while \ReMine declines less than any other methods on the rank-based measures.
In the NYT dataset, all the systems except OLLIE have similar overall precision (\ie $P@300$). But \ReMine has a ``higher'' curve since most tuples obtained by Stanford OpenIE and ClausIE will be assigned score 1. Therefore we may not rank them in a very rational way. In contrast, the scores of different tuples obtained by \ReMineG and \ReMine are usually distinct from each other. In Table~\ref{tab:performance}, \ReMine also consistently performs the best . In the Twitter dataset, \ReMine shows its power in dealing with short and noisy text. Both ClausIE and MinIE have a rather low score since there are lots of non-standard language usages and grammatical errors in tweets. In twitter, dependency parsing attaches more wrong arguments and labels than usual. All methods investigated depend on dependency parsing to varying degrees, while clause-based methods rely heavily on it and may not achieve a satisfying performance.  

\medskip
``\textit{Does global cohesiveness improve quality of open IE?}''
\smallskip

Model-wise, we believe global cohesiveness helps open IE from two aspects: (1) ranking tuples (2) updating entity phrase pairs. From Figure~\ref{fig:pr} and Table~\ref{tab:performance}, we find \ReMine outperforms \ReMineG and \ReMineL on each evaluation metric on both datasets. In particular, \ReMineG differs from \ReMineL only on extraction scores, since global cohesiveness $\sigma$ provides better ranking performance ($P@300$) over random (\ReMineL). The gain between \ReMine and \ReMineG clearly shows the updated extractions have better quality in general. 
%In the twitter dataset, a larger performance gap proves that global cohesiveness is more robust to low quality and short text compared with pattern and clause used in other methods. 

\begin{figure}[t]
	\centering
	\includegraphics[height=4cm, width=8cm]{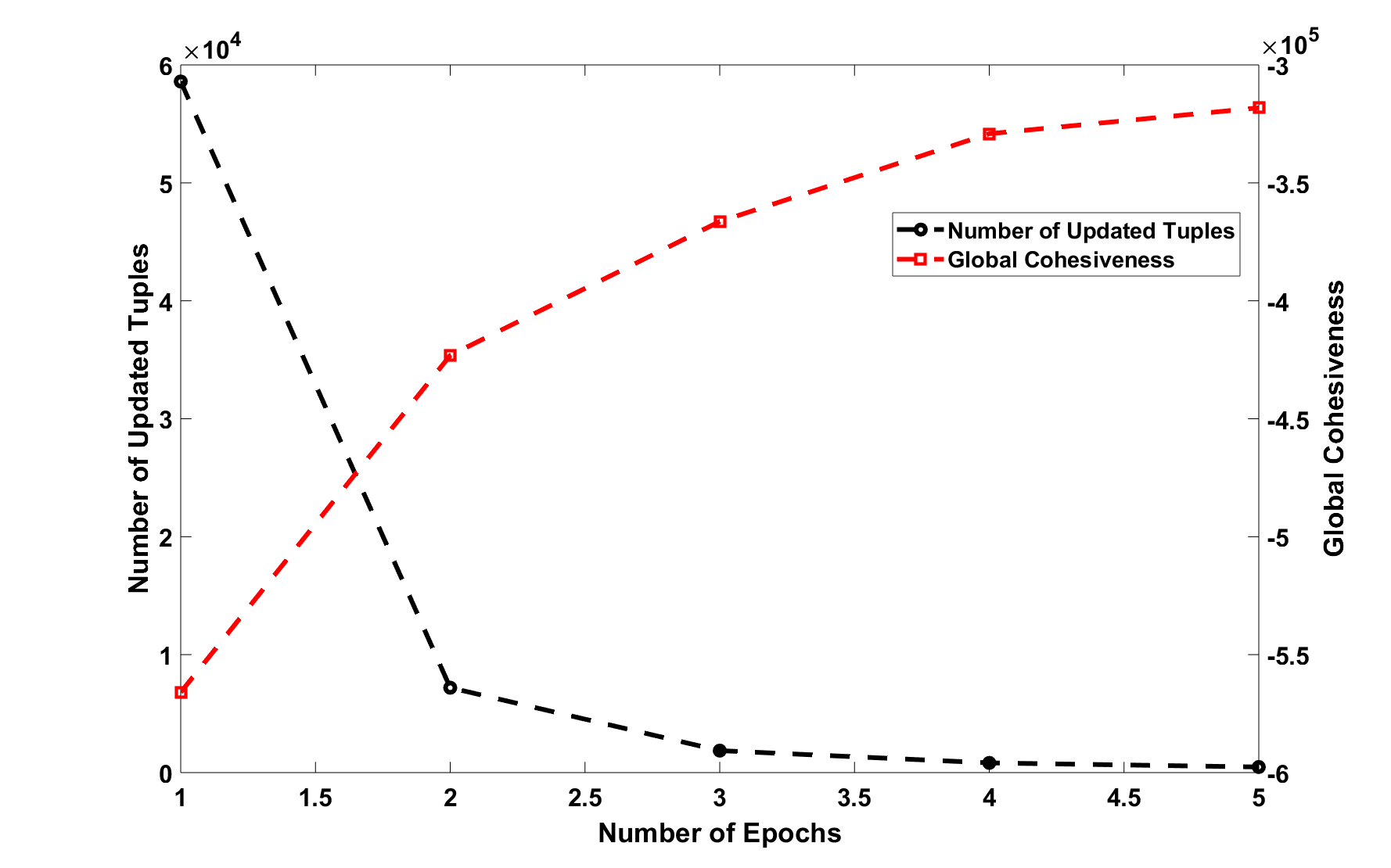}
	\vspace{-0.2cm}
	\caption{The number of updated tuples and global cohesiveness against the number of epochs for the proposed \ReMine.}
	\label{fig:ReMine-loss}
\end{figure}

\medskip
``\textit{Is the joint optimization effective and scalable?}''
\smallskip

In Fig.~\ref{fig:ReMine-loss}, We plot out the number of updated tuples and global cohesiveness objective on NYT dataset. The number of updated tuples reflects how global cohesiveness influences the tuple genration module. The convergence of global cohesiveness indicates the joint optimization leads to cleaner and more coherent extractions. 
%Overall, the joint optimization objective convergences efficiently(usually less than 10 rounds in our experiments). 
Suppose that corpus $\D$ has $N_d$ words. The time cost of phrase extraction module is $\O(\epsilon N_d)$  with the assumption that maximal length of a phrase is a constant $\epsilon$. The tuple generation module examines $M_{sp}$ candidate head entities for each entity phrase and takes $\epsilon M_{sp}$ time to perform tuple generation as maximal semantic path is bounded by $M_{sp}$. In total, it takes $\O(\epsilon M_{sp}^2 N_d )$ time. The global cohesiveness module requires $\O(N_rk + N_ek)$, where $N_r$,$N_e$ are number of entity and relation phrases and k is the embedding dimension. $N_r$ and $N_e$ is bounded by $N_d$. 
By omitting constants, the computational complexity of joint optimization is $\O(N_d)$. Furthermore, each component of \ReMine is paralleled as the independence between each document.
%!TEX root = main.tex
\section{Case Study}
%\xhdr{Massive Corpus Knowledge Discovery.} Think about this later.
%Our studies reveal an overall superior quality of extractions compared with other open IE systems and effectiveness of specific component.

\begin{table}[t]
\begin{small}
	\centering
		\caption{Extraction samples of one sentence in the NYT dataset using different methods. ``T'' means correct tuples and ``F'' means incorrect ones. $^*$The tuple is too complicated to clearly explain one proposition. $^\#$The tuple cannot read smoothly. $^\dagger$The tuple is logically wrong.}
		\vspace{-1em}
		\label{my-label}
		\begin{tabular}{p{0.2cm}p{7.1cm}p{0.1cm}}
			\hline
			\multicolumn{3}{c}{ClausIE} \\
			\hline
			$R_1$ & \textit{("Gov. Tim Pawlenty of Minnesota", "ordered", "the state health department this month to monitor day-to-day operations after state inspectors found that three men had died there in the previous month because of neglect or medical errors")} &         F$^*$                  \\
			$R_2$ & \textit{("Gov. Tim Pawlenty of Minnesota", "ordered", "the state health department this month to monitor day-to-day operations")}                                                                                                                       &             T              \\ \hline
			\multicolumn{3}{c}{Stanford OpenIE} \\ \hline
			$R_3$       & \textit{("Gov. Tim Pawlenty", "ordered", "state health department")}                                                                                                                                                                                                     &            T               \\
			$R_4$      & \textit{("Gov. Tim Pawlenty", "monitor", "operations")}                                                                                                                                                                                                 &            F$^\dagger$               \\
			$R_5$      & \textit{("three men", "died there because of", "neglect")}                                                                                                                                                                                              &           T                \\
			$R_{6}$     & \textit{("men", "died in", "month")}                                                                                                                                                                                                                    &           F$^\#$                \\ \hline
			\multicolumn{3}{c}{OLLIE} \\ \hline
			$R_{7}$     & \textit{("Gov. Tim Pawlenty of Minnesota", "ordered the state health department in", "this month")}                                                                                                                                     &          T                 \\
			$R_{8}$       & \textit{("three men", "had died there in", "the previous month")}                                                                                                                                                                                       &           T                \\ 
			$R_{9}$       & \textit{("Gov. Tim Pawlenty of Minnesota", "had died because of", "neglect errors")}                                                                                                                                                                                       &           F$^\dagger$  \\ \hline
			\multicolumn{3}{c}{MinIE} \\ \hline
			$R_{10}$     & \textit{("Tim Pawlenty", "is", "Gov.")}                                                                                                                                     &          T                 \\
			$R_{11}$       & \textit{("Tim Pawlenty of Minnesota", "ordered state health department", "this month")}                         &           T                \\
			$R_{12}$       & \textit{("QUANT\_S\_1 men", "had died because of", "neglect errors")}                                                                                                                                                                                       &           F$^\dagger$ \\ \hline
			\multicolumn{3}{c}{\ReMine} \\ \hline
			$R_{13}$       & \textit{("Gov. Tim Pawlenty of Minnesota", "order", "the state health department")}                                                                                                                                                    &           T                \\
			$R_{14}$     & \textit{("Gov. Tim Pawlenty of Minnesota", "order\_to\_monitor", "day-to-day operation")}                                                                                                                                                             &           T                \\
			$R_{15}$     & \textit{("Gov. Tim Pawlenty of Minnesota", "order\_to\_monitor\_at", "Minneapolis Veterans Home")}                                                                                                                                                                                           &           T                \\
			$R_{16}$     & \textit{("three man", "have\_die\_there", "medical error")}                                                                                                                                                                                           &           F$^\#$                \\  \hline 
	\end{tabular}
	\label{tab:case}
	\end{small}
\vspace{-0.2em} 
\end{table}

\xhdr{Clearness and correctness on extractions.}
In Table.~\ref{tab:case}, we show the extraction samples of the NYT sentence \textit{``Gov. Tim Pawlenty of Minnesota ordered the state health department this month to monitor day-to-day operations at the Minneapolis Veterans Home after state inspectors found that three men had died there in the previous month because of neglect or medical errors.''}. We can see that all the extractors share consensus on that \textit{``Gov. Tim Pawlenty of Minnesota ordered the state health department''} ($R_2, R_3, R_7, R_{11}$ and $R_{13}$). But some other actions do not belong to ``Tim Pawlenty". Both Stanford OpenIE and OLLIE make mistakes on that ($R_4$ and $R_9$). In contrast, ClausIE has no logical mistakes in the samples. However, the objective component of $R_1$ is too complicated to illustrate one proposition clearly. As we mentioned above, these kinds of tuples will be labeled as incorrect ones. 
% We also observe that all the methods except Stanford OpenIE extract \textit{``Gov. Tim Pawlenty of Minnesota''} as a phrase entity, which is a more informative phrase in contrast with \textit{``Gov. Tim Pawlenty''}. Although this difference will not affect the correctness of $R_9$, when it comes to the difference between \textit{``previous month''} and \textit{``month''}, the result will be affected (see $R_5$ and $R_{10}$). This fact also reflects the importance of detecting high quality phrases. 
$R_{15}$ is the only correct tuple to identify the location \textit{``Minneapolis Veterans Home''}, and \ReMine also carefully selects the words to form the predicate \textit{``order\_to\_monitor\_at''} to prevent excessively long relation phrase.

%\vspace{-0.2cm}
\begin{figure}[th!]
	\centering
	\begin{subfigure}[t]{0.495\linewidth}
	\includegraphics[width=\linewidth]{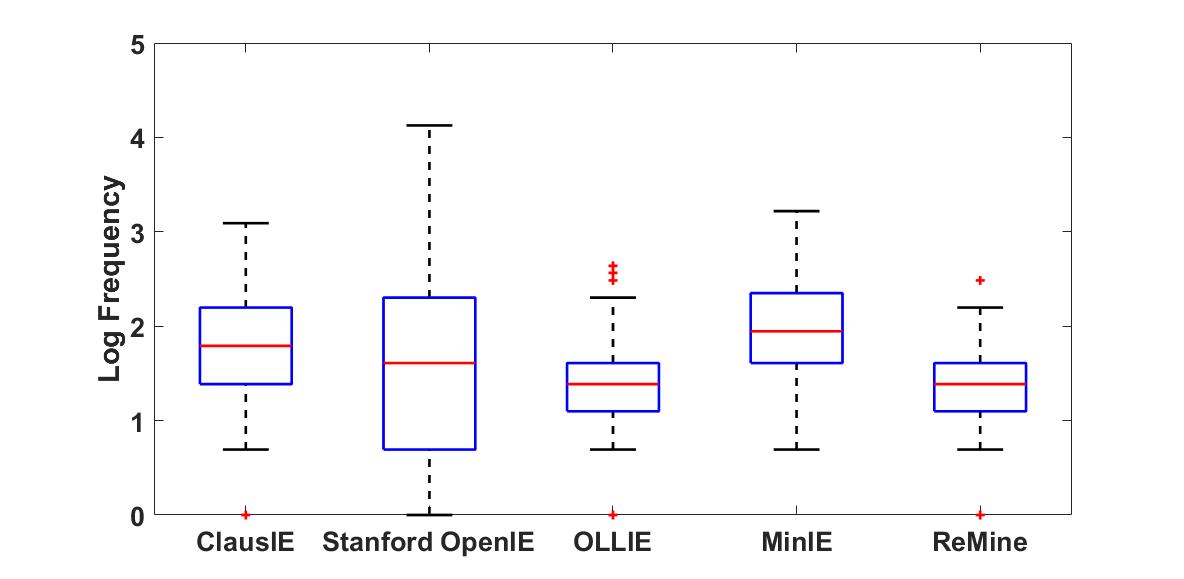}
	\caption{Number of tuples}
	\label{fig:freq_comparison}
	\end{subfigure}
	\begin{subfigure}[t]{0.495\linewidth}
	\includegraphics[width=\linewidth]{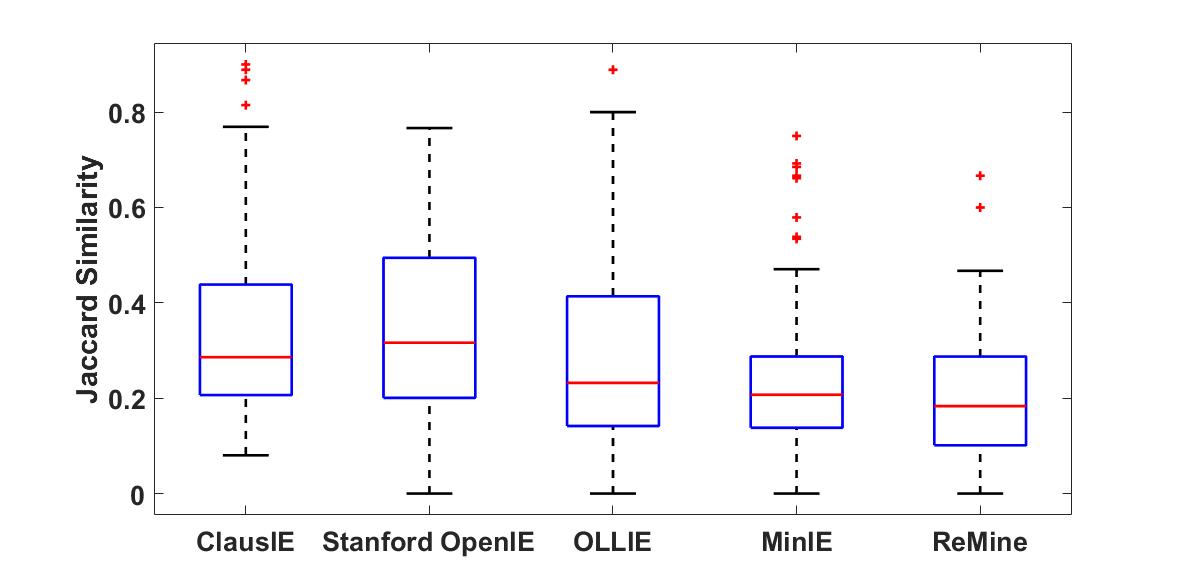}
	\caption{Jaccard similarity}
	\label{fig:sim_comparison}
	\end{subfigure}
\vspace{-1em}
	\caption{Distribution over number of extractions and distinctiveness of extractions for different Open IE systems.}
	\label{fig:distinction}
\vspace{-0.2cm}
\end{figure}

\xhdr{Distinctiveness of tuple generation.}
\label{sec:analysis_distinct}
In our formulation, we try to cover every entity detected in the target sentence while avoiding extracting duplicate tuples.
In Fig.~\ref{fig:freq_comparison}, we show the distribution of the number of extractions obtained by each Open IE system on the first 100 sentences in NYT dataset. We can see that OLLIE's and \ReMine's distributions are relatively balanced. In contrast, Stanford OpenIE returns extractions with a large variance. Among 1054 tuples it extracted, there are 228 tuples belonging to a single sentence and 157 belonging to another. This is despite the latter sentence only containing 39 words. This reminds us that the number of extractions may not be a good alternative to ``recall''. A more direct way to examine distinctiveness is calculating average Jaccard similarity between extractions from same sentence. We present the Jaccard similarity distribution of different systems at Fig.~\ref{fig:sim_comparison}, we can clearly see MinIE and \ReMine extract the most distinctive facts. 
%\subsection{Example extractions on Twitter}
%On the 2 Twitter sentences, all the extractors share the consensus on that \textit{``Madonna would have a sip of champaign for me''} ($R_{21}, R_{26}, R_{29}$ and $R_{33}$). For tweets, ClausIE's problems are no longer complexity and redundancy. The excessively complicated objective component directly makes the tuple incorrect. Due to the non-standard language usages of tweets, it becomes more difficult for Stanford Open IE and OLLIE to follow the logic of the original sentences. In contrast, \ReMine successfully extracted the main structure of the two sentences.

\xhdr{Effectiveness of global evidence.}
Corpus-level cohesiveness can help reduce local error while generating relation tuples. Especially on the twitter dataset, local linguistic structure fails to attach correct argument initially whereby global cohesiveness module corrects those extractions. In table~\ref{tab:case-global}, \ReMine rejects entity pair (\textit{Liberador}, \textit{Hollywood}) which is not compatible with the predicate ``@''. This is because in the twitter corpus, it is more common to see \textit{Person @ Place}. Therefore \ReMine attaches Hollywood to Dudamel.

%Comparing our approach with its variant (\ReMine-R), we see that by considering global cohesiveness measure, \ReMine achieves higher P@200 and MRR by ranking the same set of extractions in Fig.~\ref{fig:case-global}. We also report ranking performances of the global cohesiveness module, which is not sensitive to the choice of margin $\gamma$.

\begin{table}[th!]
\small
\caption{Different entity pairs discovered by \ReMine and \ReMineG, where blue ones are incorrect extractions.}
\vspace{-0.1cm}
\textit{Dudamel conduct his score from Liberador\#BeastMode @Hollywood Bowl}
\begin{tabular}{c c}
%\hline
	\ReMineG & \ReMine \\ %\hline
	\textit{(Dudamel; ``conduct''; Liberador)} & \textit{(Dudamel; ``conduct'';	Liberador)} \\
	\textit{(\color{blue}{Dudamel; ``conduct...from'';}}  & \textit{(Dudamel; ``conduct... @'';}  \\
	\textit{\color{blue}{\#BeastMode})} & \textit{Hollywood Bowl)} \\
	\textit{(\color{blue}{Liberador, ``@'', Hollywood Bowl})} & \\
%\hline
\end{tabular}
%\small
\label{tab:case-global} 

\end{table}

%\xhdr{Scalability.}
%!TEX root = main.tex
\section{RELATED WORK}
\xhdr{Open Information Extraction.} Open domain information extraction has been extensively studied in literature. 
Most of the existing work follow two lines of work, that is, pattern based methods or clause based methods. 
Pattern based information extraction can be as early as Hearst patterns like ``$NP_0$ such as \{$NP_1, NP_2, ... $\}'' for hyponymy relation extraction~\cite{hearst1992automatic}. 
Carlson and Mitchell \etal introduced Never-Ending Language Learning (NELL) based on free-text predicate patterns \cite{carlson2010toward,mitchell2015never}. 
ReVerb~\cite{fader2011identifying} identified relational phrases via part-of-speech-based regular expressions. 
Besides part-of-speech tags, recent works have started to use more linguistic features, such as dependency parsing, to induce long distance relationships~\cite{nakashole2012patty,schmitz2012open}. 
Similarly, ClausIE~\cite{del2013clausie} inducted short but coherent pieces of information along dependency paths, which is typically subject, predicate and optional object with complement. 
Angeli \etal adopts a clause splitter using distant training and statistically maps predicate to known relation schemas~\cite{angeli2015leveraging}. MinIE~\cite{gashteovski2017minie} removes overly-specific constituents and captures implicit relations in ClausIE by introducing several statistical measures like polarity, modality, attribution, and quantities.
 Compared with these works, this paper differs in several aspects: (1) previous work relies on external tools for phrase extraction, which may suffer from domain-shift and sparsity problem, while we provide an End-to-End solution towards Open IE. (2) Although previous efforts achieve comparable high precision and reasonable coverage on extraction results, they all focus on local linguistic context. The correctness of extracted facts are evaluated purely on local context, however, large corpus can exclude false extractions from inferred inconsistencies.

\xhdr{Knowledge Base Embedding and Completion.} Knowledge bases (KBs), such as DBpedia~\cite{bollacker2008freebase} and Freebase~\cite{lehmann2015dbpedia}, extract tuples from World Wide Web. Knowledge base population or completion aims at predicting whether tuples not in knowledge base are likely to be true or not. Previous works attempted to construct web-scale knowledge base using statistical learning and pre-defined rules and predicates~\cite{niu2012deepdive}.
Recently, embedding models~\cite{bordes2013translating, lin2015learning, socher2013reasoning, rocktaschel2015injecting} have been widely used to learn semantic representation for both entities and relations. 
By observing each relation may have different semantic meaning, Wang~\etal~\cite{wang2014knowledge} projected entity vectors to relation-specific hyperplane. 
Further research~\cite{guu2015traversing,luo2015context} shows that embedding techniques can support composite query(\ie asking about multiple relations) on knowledge graph. All previouos knowledge graph embedding methods start with existing knowledge base tuples, while our proposed global cohesiveness representation starts from noisy extractions.
There is another line of work trying to combining KB relations and textual relations~\cite{toutanova2015representing} or model unstructured and structured data by universal schema~\cite{riedel2013relation}. However, they are all built upon on existing and specific relation types.
Although we shared similar semantic measures as these work, \ReMine uses KB embeddings to measure quality of extracted relation tuples and improve Open IE in a multi-tasking way. 
%But their main contribution is improving knowledge base completion while our goal is producing unstructured relation tuples and organizing them into a domain-specific knowledge graph.

%On the opposite, we output comparable clean relation tuples rather than taking tuples as input.

%\textbf{Phrase Mining}
%Data-driven phrase extraction do not rely on domain-specific linguistic rules or exhausted human labeling. Instead, phrase mining work~\cite{liu2015mining,shang2017automated,el2014scalable} use frequency statistics to both approach n-gram candidates and quality estimation. Shang \etal~\cite{shang2017automated} proposed a method, that can purely use distant training data to train its phrase quality estimator. Regarding the tradeoff between scalability and accuracy, previous methods deploy phrasal segmentation in a context-free manner. It is still quite challenging that how to efficiently include out-of-pattern information.
%\QZ{Jingbo, do you think related work about Phrase Mining looks good?, do we need to introduce more related work as AutoPhrase}.  
%!TEX root = main.tex
\section{Conclusion}
This paper studies the task of open information extraction and proposes a principled framework, \ReMine, to unify local contextual information and global structural cohesiveness for effective extraction of relation tuples. 
%\ReMine leverages distant supervision in conjunction with existing knowledge bases to provide automatically-labeled sentence and guide the entity and relation segmentation. 
The local objective is jointly learned together with a translating-based objective to enforce structural cohesiveness, such that corpus-level statistics are incorporated for boosting high-quality tuples extracted from individual sentences. Experiments on two real-world corpora of different domains demonstrate that \ReMine system achieves superior precision when outputting same number of extractions, compared with several state-of-the-art open IE systems. Interesting future work can be (1) On-The-Fly knowledge graph construction from relation tuples; (2) applying \ReMine to downstream applications \eg open domain Question Answering.
\section{Acknowledgements}
Research was sponsored in part by the U.S. Army Research Lab.
under Cooperative Agreement No. W911NF-09-2-0053 (NSCTA),
National Science Foundation IIS 16-18481, IIS 17-04532, and IIS-17-
41317, and grant 1U54GM114838 awarded by NIGMS through funds
provided by the trans-NIH Big Data to Knowledge (BD2K) initiative
(www.bd2k.nih.gov). Xiang Ren's research has been supported in part by National Science Foundation SMA 18-29268. We thank Frank F. Xu and Ellen Wu for valuable feedback and discussions.

%\scriptsize
\bibliography{ref,cotype}

\begin{thebibliography}{10}

\bibitem{allahverdyan2011comparative}
A.~Allahverdyan and A.~Galstyan.
\newblock Comparative analysis of viterbi training and maximum likelihood
  estimation for hmms.
\newblock In {\em NIPS}, 2011.

\bibitem{angeli2015leveraging}
G.~Angeli, M.~J. Premkumar, and C.~D. Manning.
\newblock Leveraging linguistic structure for open domain information
  extraction.
\newblock In {\em ACL}, 2015.

\bibitem{banko2007open}
M.~Banko, M.~J. Cafarella, S.~Soderland, M.~Broadhead, and O.~Etzioni.
\newblock Open information extraction from the web.
\newblock In {\em IJCAI}, 2007.

\bibitem{bollacker2008freebase}
K.~Bollacker, C.~Evans, P.~Paritosh, T.~Sturge, and J.~Taylor.
\newblock Freebase: a collaboratively created graph database for structuring
  human knowledge.
\newblock In {\em SIGMOD}, 2008.

\bibitem{bordes2013translating}
A.~Bordes, N.~Usunier, A.~Garcia-Duran, J.~Weston, and O.~Yakhnenko.
\newblock Translating embeddings for modeling multi-relational data.
\newblock In {\em NIPS}, 2013.

\bibitem{Bunescu2007LearningTE}
R.~C. Bunescu and R.~J. Mooney.
\newblock Learning to extract relations from the web using minimal supervision.
\newblock In {\em ACL}, 2007.

\bibitem{carlson2010toward}
A.~Carlson, J.~Betteridge, B.~Kisiel, B.~Settles, E.~R. Hruschka~Jr, and T.~M.
  Mitchell.
\newblock Toward an architecture for never-ending language learning.
\newblock In {\em AAAI}, 2010.

\bibitem{carlson2010coupled}
A.~Carlson, J.~Betteridge, R.~C. Wang, E.~R. Hruschka~Jr, and T.~M. Mitchell.
\newblock Coupled semi-supervised learning for information extraction.
\newblock In {\em WSDM}, 2010.

\bibitem{isem2013daiber}
J.~Daiber, M.~Jakob, C.~Hokamp, and P.~N. Mendes.
\newblock Improving efficiency and accuracy in multilingual entity extraction.
\newblock In {\em I-Semantics}, 2013.

\bibitem{del2013clausie}
L.~Del~Corro and R.~Gemulla.
\newblock Clausie: clause-based open information extraction.
\newblock In {\em WWW}, 2013.

\bibitem{dong2014knowledge}
X.~L. Dong, T.~Strohmann, S.~Sun, and W.~Zhang.
\newblock Knowledge vault: A web-scale approach to probabilistic knowledge
  fusion.
\newblock In {\em KDD}, 2014.

\bibitem{fader2011identifying}
A.~Fader, S.~Soderland, and O.~Etzioni.
\newblock Identifying relations for open information extraction.
\newblock In {\em EMNLP}, 2011.

\bibitem{faderopenQA14}
A.~Fader, L.~Zettlemoyer, and O.~Etzioni.
\newblock Open question answering over curated and extracted knowledge bases.
\newblock {\em KDD}, 2014.

\bibitem{gashteovski2017minie}
K.~Gashteovski, R.~Gemulla, and L.~Del~Corro.
\newblock Minie: minimizing facts in open information extraction.
\newblock In {\em EMNLP}, 2017.

\bibitem{guu2015traversing}
K.~Guu, J.~Miller, and P.~Liang.
\newblock Traversing knowledge graphs in vector space.
\newblock In {\em EMNLP}, 2015.

\bibitem{hearst1992automatic}
M.~A. Hearst.
\newblock Automatic acquisition of hyponyms from large text corpora.
\newblock In {\em ACL}, 1992.

\bibitem{lehmann2015dbpedia}
J.~Lehmann, R.~Isele, M.~Jakob, A.~Jentzsch, D.~Kontokostas, P.~N. Mendes,
  S.~Hellmann, M.~Morsey, P.~Van~Kleef, S.~Auer, et~al.
\newblock Dbpedia--a large-scale, multilingual knowledge base extracted from
  wikipedia.
\newblock {\em Semantic Web}, 6(2):167--195, 2015.

\bibitem{lin2015learning}
Y.~Lin, Z.~Liu, M.~Sun, Y.~Liu, and X.~Zhu.
\newblock Learning entity and relation embeddings for knowledge graph
  completion.
\newblock In {\em AAAI}, volume~15, pages 2181--2187, 2015.

\bibitem{ling2012fine}
X.~Ling and D.~S. Weld.
\newblock Fine-grained entity recognition.
\newblock In {\em AAAI}, 2012.

\bibitem{book_pm}
J.~Liu, J.~Shang, and J.~Han.
\newblock Phrase mining from massive text and its applications.
\newblock volume~9.

\bibitem{liu2015mining}
J.~Liu, J.~Shang, C.~Wang, X.~Ren, and J.~Han.
\newblock Mining quality phrases from massive text corpora.
\newblock In {\em SIGMOD}, 2015.

\bibitem{2017arXiv170904109L}
L.~Liu, J.~Shang, F.~F. Xu, X.~Ren, H.~Gui, J.~Peng, and J.~Han.
\newblock Empower sequence labeling with task-aware neural language model.
\newblock {\em arXiv preprint arXiv:1709.04109}, 2017.

\bibitem{luo2015context}
Y.~Luo, Q.~Wang, B.~Wang, and L.~Guo.
\newblock Context-dependent knowledge graph embedding.
\newblock In {\em EMNLP}, 2015.

\bibitem{ma2016end}
X.~Ma and E.~Hovy.
\newblock End-to-end sequence labeling via bi-directional lstm-cnns-crf.
\newblock In {\em ACL}, 2016.

\bibitem{manning2014stanford}
C.~D. Manning, M.~Surdeanu, J.~Bauer, J.~R. Finkel, S.~Bethard, and
  D.~McClosky.
\newblock The stanford corenlp natural language processing toolkit.
\newblock In {\em ACL (System Demonstrations)}, 2014.

\bibitem{mitchell2015never}
T.~M. Mitchell, W.~W. Cohen, E.~R. Hruschka~Jr, P.~P. Talukdar, J.~Betteridge,
  A.~Carlson, B.~D. Mishra, M.~Gardner, B.~Kisiel, J.~Krishnamurthy, et~al.
\newblock Never ending learning.
\newblock In {\em AAAI}, 2015.

\bibitem{nakashole2012patty}
N.~Nakashole, G.~Weikum, and F.~Suchanek.
\newblock Patty: A taxonomy of relational patterns with semantic types.
\newblock In {\em EMNLP-CoNLL}, 2012.

\bibitem{niu2012deepdive}
F.~Niu, C.~Zhang, C.~R{\'e}, and J.~W. Shavlik.
\newblock Deepdive: Web-scale knowledge-base construction using statistical
  learning and inference.
\newblock In {\em VLDS}, 2012.

\bibitem{riedel2013relation}
S.~Riedel, L.~Yao, A.~McCallum, and B.~M. Marlin.
\newblock Relation extraction with matrix factorization and universal schemas.
\newblock In {\em NAACL}, 2013.

\bibitem{rocktaschel2015injecting}
T.~Rockt{\"a}schel, S.~Singh, and S.~Riedel.
\newblock Injecting logical background knowledge into embeddings for relation
  extraction.
\newblock In {\em Proceedings of the 2015 Conference of the North American
  Chapter of the Association for Computational Linguistics: Human Language
  Technologies}, pages 1119--1129, 2015.

\bibitem{sarawagi2008information}
S.~Sarawagi.
\newblock Information extraction.
\newblock {\em Foundations and trends in databases}, 1(3):261--377, 2008.

\bibitem{schmitz2012open}
M.~Schmitz, R.~Bart, S.~Soderland, O.~Etzioni, et~al.
\newblock Open language learning for information extraction.
\newblock In {\em EMNLP-CoNLL}, 2012.

\bibitem{shang2017automated}
J.~Shang, J.~Liu, M.~Jiang, X.~Ren, C.~R. Voss, and J.~Han.
\newblock Automated phrase mining from massive text corpora.
\newblock {\em arXiv preprint arXiv:1702.04457}, 2017.

\bibitem{socher2013reasoning}
R.~Socher, D.~Chen, C.~D. Manning, and A.~Ng.
\newblock Reasoning with neural tensor networks for knowledge base completion.
\newblock In {\em Advances in neural information processing systems}, pages
  926--934, 2013.

\bibitem{sun2015open}
H.~Sun, H.~Ma, W.-t. Yih, C.-T. Tsai, J.~Liu, and M.-W. Chang.
\newblock Open domain question answering via semantic enrichment.
\newblock In {\em WWW}, 2015.

\bibitem{toutanova2015representing}
K.~Toutanova, D.~Chen, P.~Pantel, H.~Poon, P.~Choudhury, and M.~Gamon.
\newblock Representing text for joint embedding of text and knowledge bases.
\newblock In {\em Proceedings of the 2015 Conference on Empirical Methods in
  Natural Language Processing}, pages 1499--1509, 2015.

\bibitem{wang2014knowledge}
Z.~Wang, J.~Zhang, J.~Feng, and Z.~Chen.
\newblock Knowledge graph embedding by translating on hyperplanes.
\newblock In {\em AAAI}, 2014.

\bibitem{zhang2016geoburst}
C.~Zhang, G.~Zhou, Q.~Yuan, H.~Zhuang, Y.~Zheng, L.~Kaplan, S.~Wang, and
  J.~Han.
\newblock Geoburst: Real-time local event detection in geo-tagged tweet
  streams.
\newblock In {\em SIGIR}, 2016.

\end{thebibliography}
\bibliographystyle{abbrv}

\end{document}